\documentclass[10pt,twocolumn,letterpaper]{article}

\usepackage{cvpr}              

\usepackage{amssymb}
\usepackage{pifont}
\newcommand{\cmark}{\ding{51}}%
\newcommand{\xmark}{\ding{55}}%

\newcommand{\walon}[1]{\textcolor{black}{#1}}
\newcommand{\bor}[1]{\textcolor{black}{#1}}

\usepackage[dvipsnames]{xcolor}

\usepackage{multirow}
\usepackage{graphicx}
\usepackage{amsmath}
\usepackage{tabularray}
\usepackage{float}

\usepackage{amsmath,amsfonts,bm}

\def\eqref#1{equation~\ref{#1}}

\def\1{\bm{1}}

\DeclareMathAlphabet{\mathsfit}{\encodingdefault}{\sfdefault}{m}{sl}
\SetMathAlphabet{\mathsfit}{bold}{\encodingdefault}{\sfdefault}{bx}{n}

\usepackage{indentfirst}
\usepackage{tabularx}
\usepackage{dsfont}
\usepackage[linesnumbered,ruled,vlined]{algorithm2e}

\usepackage{subcaption}
\usepackage{footmisc}
\usepackage[accsupp]{axessibility}

\definecolor{cvprblue}{rgb}{0.21,0.49,0.74}
\usepackage[pagebackref,breaklinks,colorlinks,citecolor=cvprblue]{hyperref}

\def\paperID{142} 
\def\confName{CVPR}
\def\confYear{2024}

\begin{document}
\title{MCPNet: An Interpretable Classifier via Multi-Level Concept Prototypes}

\author{%
  Bor-Shiun Wang\textsuperscript{$\dagger$} 
  \quad Chien-Yi Wang\textsuperscript{$^* \ddagger$}
  \quad Wei-Chen Chiu\textsuperscript{$^* \dagger$}\\
  \textsuperscript{$\dagger$}National Yang Ming
Chiao Tung University \quad \textsuperscript{$\ddagger$}NVIDIA Research\\  
  \small{\texttt{eddiewang.cs10@nycu.edu.tw}}, \;  
  \small{\texttt{chienyiw@nvidia.com}}, \; 
  \small{\texttt{walon@cs.nctu.edu.tw}}
}


\twocolumn[{%
\renewcommand\twocolumn[1][]{#1}%
\maketitle
\vspace{-14mm}

\begin{minipage}{\textwidth}
\begin{figure}[H]\centering
    \centering
    \includegraphics[width=0.99\textwidth]{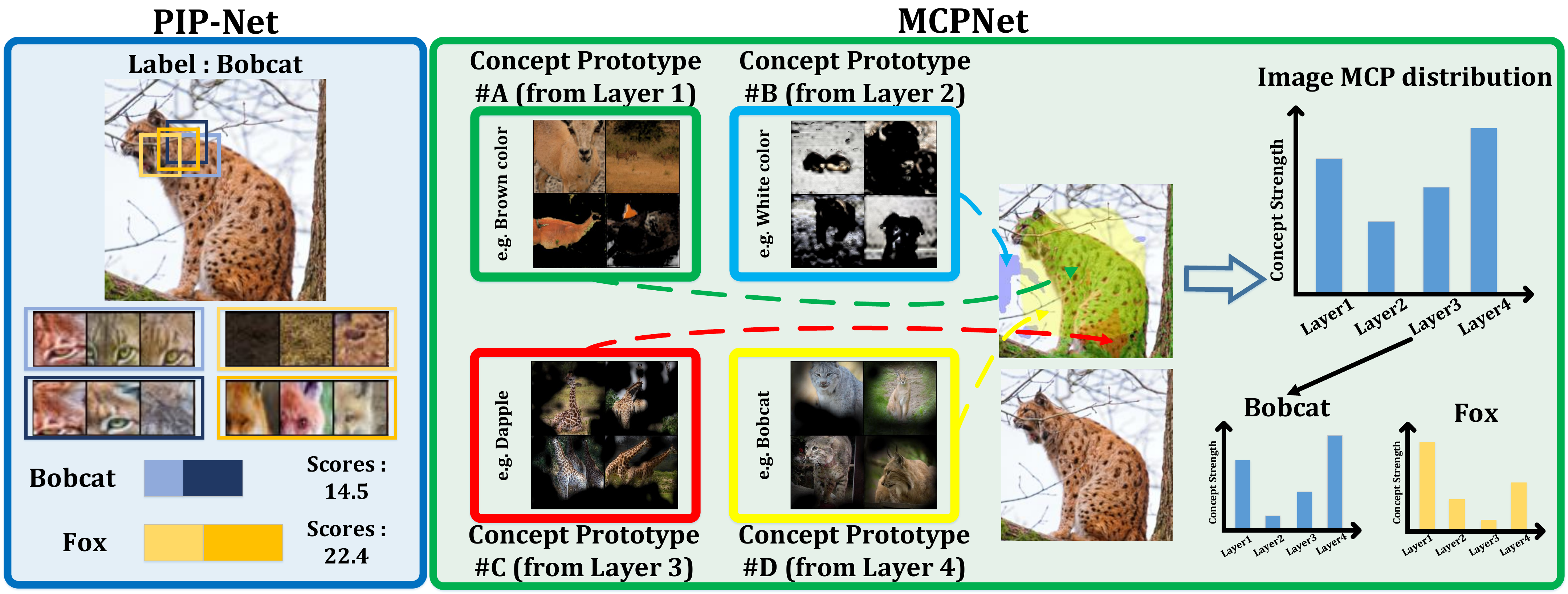}
    \vspace{-0.3cm}
    \caption{\textbf{The “Bobcat” was correctly classified by our MCPNet but incorrectly classified as a “Fox” by PIP-Net~\cite{nauta2023pip}.} On the right side, we provide an illustration of using our proposed Multi-level Concept Prototype (MCP) distribution to classify and explain the input image. In particular, our concept prototypes are extracted from multiple layers of the classification model (thus having low-level to high-level concepts). In comparison with a recent state-of-the-art baseline, PIP-Net~\cite{nauta2023pip} shown on the left side which only adopts single-level explanations (symbolized as colorful boxes on the bottom portion, they are usually extracted from the last model layer), our proposed MCPNet provides more comprehensive explanations as well as better classification performance.}
    \vspace{0.2cm}
    \label{fig:teaser_figure}
\end{figure}
\end{minipage}
}]{
  \renewcommand{\thefootnote}%
    {\fnsymbol{footnote}}
  \footnotetext[1]{Equal advising}
}


\begin{abstract}
\vspace{-0.3cm}
Recent advancements in post-hoc and inherently interpretable methods have markedly enhanced the explanations of black box classifier models. These methods operate either through post-analysis or by integrating concept learning during model training. Although being effective in bridging the semantic gap between a model's latent space and human interpretation, these explanation methods only partially reveal the model's decision-making process. The outcome is typically limited to high-level semantics derived from the last feature map. We argue that the explanations lacking insights into the decision processes at low and mid-level features are neither fully faithful nor useful. Addressing this gap, we introduce the Multi-Level Concept Prototypes Classifier (MCPNet), an inherently interpretable model. MCPNet autonomously learns meaningful concept prototypes across multiple feature map levels using Centered Kernel Alignment (CKA) loss and an energy-based weighted PCA mechanism, and it does so without reliance on predefined concept labels. \bor{Further, we propose a novel classifier paradigm that learns and aligns multi-level concept prototype distributions for classification purposes via Class-aware Concept Distribution (CCD) loss.} Our experiments reveal that our proposed MCPNet while being adaptable to various model architectures, offers comprehensive multi-level explanations while maintaining classification accuracy. Additionally, its concept distribution-based classification approach shows improved generalization capabilities in few-shot classification scenarios. 
Project page is available \url{https://eddie221.github.io/MCPNet/}.
\vspace{-0.5cm}

\end{abstract}    
\section{Introduction}

\begin{table*}[ht]
\centering
\resizebox{\textwidth}{!}{
\begin{tabular}{l|c|c|c|c|c}
\toprule
 & \textbf{MCPNet (Ours)} & ProtoPNet~\cite{chen2019looks, donnelly2022deformable, nauta2023pip} & Concept Bottleneck \cite{koh2020concept} & TCAV~\cite{kim2018interpretability} & CRAFT \cite{fel2023craft} \\ 
\midrule
Explanation Type & Inherently & Inherently & Inherently & Post-hoc & Post-hoc \\
Explanation Scale & Multi-Level & Single-Level & Single-Level & Single-Level & Single-Level \\
w/o Concept Labels & \cmark & \cmark & \xmark & \xmark* & \cmark \\
w/o Modifying Models & \cmark & \xmark & \xmark & \cmark & \cmark** \\
\bottomrule
\end{tabular}
}
\vspace{-0.3cm}
\caption{We compare our method with four distinct lines of explainable approaches. Our MCPNet inherently offers multi-level explanations without the need for model modifications or concept labels, making it competitive compared to methods that achieve only partial properties. *It's noteworthy that TCAV requires dataset preparation both with and without specific concepts. **Only for non-negative features due to the limitation of Non-Negative Matrix Factorization.} 
\vspace{-0.4cm}
\label{tab:Comparison_method}
\end{table*}
The rapid integration of deep learning across various domains has brought to the forefront a critical question: what underlying mechanisms drive the decisions of these models? This query has led to the emergence of Explainable Artificial Intelligence (XAI), a field dedicated to demystifying the operations of opaque 'black box' models.

In XAI, numerous methods have been developed, addressing different aspects of model interpretability~\cite{ribeiro2016should, lundberg2017unified, kim2018interpretability, zhang2019interpreting, koh2020concept, gu2021semantics, wang2023learning, nauta2023pip, kim2023grounding, fel2023craft}. These methods generally fall into two categories: 1) post-hoc methods, and 2) inherently interpretive methods. Post-hoc methods focus on elucidating model behaviours either locally~\cite{ribeiro2016should} or globally~\cite{lundberg2017unified}, offering explanations for predictions without necessitating model retraining. While being valuable, these methods often provide explanations that lack coherence with the models' decision-making processes, leading to potential issues of unfaithfulness in interpretation~\cite{rudin2019stop}.

To address the limitations of unfaithfulness in post-hoc methods, there has been a growing emphasis on inherently interpretable models featuring built-in, case-based reasoning processes. In contrast to post-hoc methods, these models generate explanations that are integral to the classification process. As one of the pioneering work, Concept Bottleneck Model (CBM)~\cite{koh2020concept} which first translates an image into features signifying the presence or absence of predefined concepts, and then bases its decision-making on these conceptual representations. Recognizing the challenge of acquiring pre-defined concept labels, subsequent research has shifted towards the autonomous identification of concepts during training. Notable among these are ProtoPNet~\cite{chen2019looks} and its derivatives~\cite{donnelly2022deformable, nauta2021neural, rymarczyk2020protopshare, rymarczyk2022interpretable, wang2021interpretable, wang2023learningsupport, nauta2023pip}, which build inherently interpretable classifier models with a predetermined number of prototypical (concept) parts that are learned automatically during the training process.

While recent advancements in inherently interpretable methods have significantly improved explanations of black box classifier models, a common limitation persists: these methods typically derive explanations from a single part of the model. Most of the previous studies (e.g., ProtoPNet series) have concentrated on extracting human-understandable concepts from the last feature map, just before the fully connected (FC) layer, to elucidate model behavior (noting that here we take the classification models as the representative example, without loss of generality). This line of approaches transform features into explanations to inform outcomes, yielding understandable concepts but only illuminating the last model layer with high-level semantics, leaving much still obscured as a 'black box'.

In this paper, we present MCPNet, a novel \textbf{hierarchical explainable classifier} designed for more comprehensive multi-level model explanations. By eliminating the FC layer, MCPNet encourages learning of more distinctive features across various model layers. Unlike previous methods that used entire channels to represent concept features, often facing challenges in establishing orthogonality among similar concepts \cite{doumanoglou2023unsupervised}, MCPNet partitions feature maps into distinct \textbf{segments}. Each segment focuses on learning unique concepts, facilitated by our proposed Centered Kernel Alignment (CKA) loss. These segments are further differentiated using a weighted Principal Component Analysis (PCA), which prioritizes pixel importance in extracting the concept prototype (CP). For each image, MCPNet calculates the concept response using the CP and its corresponding concept segment, forming what we term the Multi-level Concept Prototype distribution (MCP distribution). Additionally, we introduce the Class-aware Concept Distribution (CCD) loss. This loss function enhances the distinction of the MCP distribution between different classes while minimizing it within the same class. To classify images without the conventional FC layer, MCPNet compares the image's MCP distribution with the class-specific centroid MCP distribution, which is an average of the MCP distributions across instances in the same category, to identify the most similar class. The primary contributions of our work are outlined as follows:
\vspace{-0.1cm}

\begin{itemize}
    \item We introduce a novel hierarchical explainable classifier MCPNet that offers in-depth, multi-level explanations of model behavior. This advancement marks a significant shift from traditional models, which primarily focus on high-level semantics, to a more comprehensive approach that includes insights from various layers of the model.

    \item The proposed inherently multi-level interpretable paradigm can be seamlessly integrated with multiple convolution-based model architectures without additional modules or trainable parameters while maintaining comparable classification performance. 

    \item Evaluation on several benchmark datasets verifies that our method can provide richer concept-based explanations across low-to-high semantic levels and exhibit better generalization ability towards unseen categories.  

\end{itemize}

\section{Related Works}

\paragraph{Post-hoc Interpretation Methods}
In the realm of post-hoc explanations, a variety of methods have been developed to elucidate model behaviors without necessitating retraining. These approaches predominantly utilize extracted features from specific instances. A notable category within this field is \textit{attribution methods} which are initially proposed by~\cite{simonyan2013deep} and subsequently inspiring further research~\cite{selvaraju2017grad, sundararajan2017axiomatic, smilkov2017smoothgrad}. These methods primarily generate heatmaps to highlight the impact of individual pixels on model outcomes.

Alternatively, \textit{concept-based methods} aim to derive human-understandable concepts from model features. 
By using the predefined concept, TCAV~\cite{kim2018interpretability} measures the concept impact on models' outputs.
A recent advancement in this area is CRAFT~\cite{fel2023craft}, which employs Non-Negative Matrix Factorization (NMF) to deconstruct target features and iteratively clarify ambiguous concepts from lower-layer features. Model-agnostic methods also play a pivotal role, with LIME~\cite{ribeiro2016should} introducing a local explanation technique that assesses the influence of feature presence or absence on model results through input perturbation. In contrast, SHAP~\cite{lundberg2017unified} utilizes Shapley values from game theory to provide individual prediction explanations on a global scale. Despite their utility, these methods have significant limitations, primarily in providing explanations that may not align with model predictions. This discordance raises concerns about the reliability of explanations, as it becomes challenging to discern whether inaccuracies lie in the explanation or stem from reliance on spurious data in predicting outcomes.

\vspace{-0.5cm}
\paragraph{Inherently Interpretable Methods}
Inherently interpretable methods, which learn explanations during the training process, result in outcomes more closely aligned with the model's decision-making process due to these integrated explanations being more faithful and reliable. The Concept Bottleneck Model (CBM)~\cite{koh2020concept} exemplifies this by mapping images to features that represent the presence or absence of predefined concepts, subsequently utilizing these concepts for decision-making. However, this method's reliance on predefined concept labels restricts its ability to discriminate concepts not explicitly provided in the data.

Alternatively, ProtoPNet~\cite{chen2019looks} learns class-specific prototypes, akin to concepts, with a set number per class. It classifies by calculating responses from each class's prototype and summarizing these responses using a fully connected layer. Explanations are derived as a weighted sum of all prototypes. Deformable ProtoPNet~\cite{donnelly2022deformable} enhances this approach by training the prototypes to be orthogonal, aiding in clarity of interpretation. Similarly, TesNet~\cite{wang2021interpretable} incorporates an additional module to separate prototypes on the Grassmann manifold.

To optimize the number of prototypes, ProtoPShare~\cite{rymarczyk2020protopshare} merges similar prototypes, while ProtoPool~\cite{rymarczyk2022interpretable} adopts a soft-assignment approach for prototype sharing across classes. ProtoTree~\cite{nauta2021neural} utilizes a binary decision tree, allowing prototypes to be shared across all classes, thereby reducing their number. This process resembles traditional decision trees, where image features traverse the tree, aggregating logits at leaf nodes and considering probabilities from the root to the leaves. More recent approaches, such as PIP-Net~\cite{nauta2023pip}, strive for prototypes that align more closely with human perception by bridging the gap between latent and pixel spaces. ST-ProtoPNet~\cite{wang2023learningsupport} introduces support prototypes that position near the classification boundary to enhance discriminative capabilities.

Despite these methods significantly enhancing explanations for black-box models, their focus on high-level features leaves the behaviors of low and mid-layer features largely opaque. Our proposed MCPNet addresses this gap by providing explanations via multi-level concept prototypes. \Cref{tab:Comparison_method} compares our MCPNet with four representative lines of approaches, which summarizes the difference in multiple critical aspects.

\begin{figure*}
    \centering
    \includegraphics[width=0.99\textwidth]{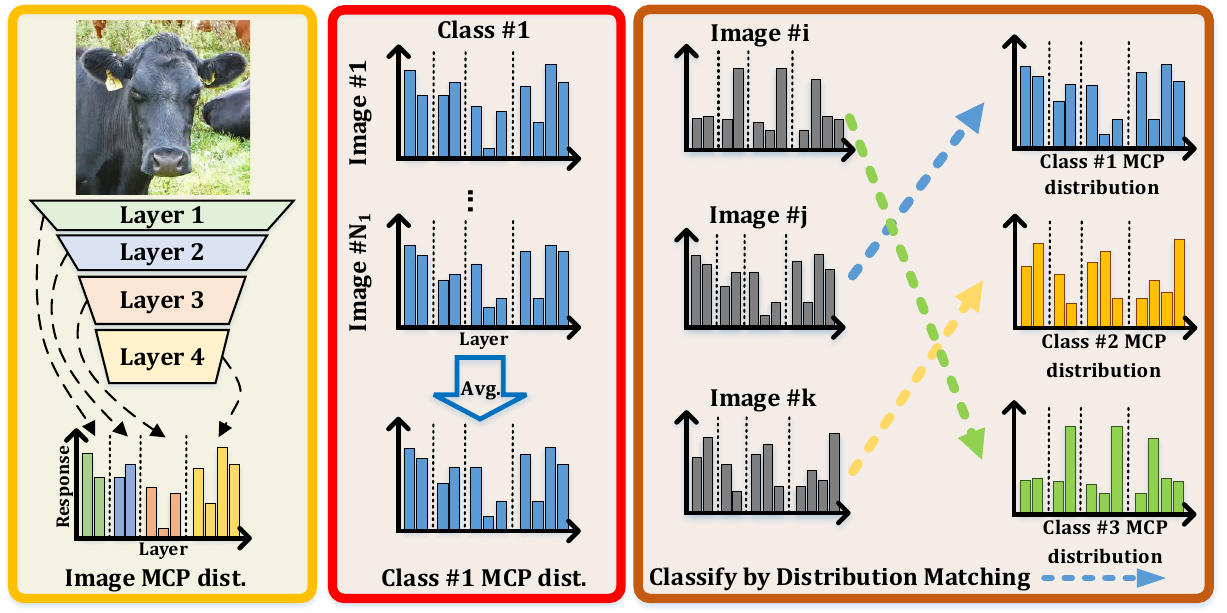}
    \vspace{-0.4cm}
    \caption{The overall workflow of our MCPNet. To classify the image, we first calculate the class-specific MCP distribution (yellow box) by averaging the MCP distributions from instances of specific classes in the training set (red box). Utilizing the class-specific MCP distribution, images are classified by identifying the most similar class via calculating the Jensen-Shannon (JS) divergence (brown box).}
    \vspace{-0.4cm}
    \label{fig:Overall_workflow}
\end{figure*}

\vspace{-0.3cm}
\section{Method}
\subsection{Overall Framework}

\walon{As motivated previously, in this paper we introduce an inherently interpretable framework named MCPNet to reveal multi-level global concepts throughout different model layers. This marks a departure from previous approaches which usually provide only single-level explanations. Moreover, our framework has the capability to classify images via adopting the distribution of multi-level concept prototypes instead of relying on the typical fully connected (FC) classifier, in which it ensures the model to consider not only the features from the last layer but also from mid- and even low-layers, thus providing better generalizability towards unseen categories. In the following we sequentially detail the main components and important designs of our proposed MCPNet framework.}

\subsection{Centered Kernel Alignment (CKA) Loss}

\walon{Given a feature map $F_l \in \mathbb{R}^{B \times C_l \times H_l \times W_l}$ of dimension (width $H_l \times$ height $W_l \times$ channels $C_l$) and batch size $B$ obtained from the $l$-th layer of a deep model, as the general architecture design of deep neural networks by nature extracts different characteristics of the input data into the features placed along the channels (i.e. each channel stands for a certain data characteristic), we now would like to partition the feature map $F_l$ along the channel dimension into several distinct segments where each segment of size $\mathbb{R}^{B \times C^\prime_l \times H_l \times W_l}$ groups $C^\prime_l$ channels to form a more semantic-meaningful component (representing a specific combination of data characteristics), termed as ``concept''. These concepts in results serve as a bridge/proxy to provide more interpretable explanations upon the input data samples towards their corresponding categories/classes. Moreover, the concepts ideally should be diverse and discriminative from each other in order to describe the data from different aspects, forming a more concise but representative basis of interpretation. To this end, we introduce Centered Kernel Alignment (CKA) loss $\mathcal{L}^{\mathbf{CKA}}$, which leverages CKA metric \cite{kornblith2019similarity} to measure the similarity among segments and its minimization brings the distinct concepts (i.e. the concepts are learned to be dissimilar and independent from each other).}

\walon{Basically, given two concept segments $\mathcal{X}$ and $\mathcal{Y}$, their CKA similarity $\mathbf{CKA}(\mathcal{X}, \mathcal{Y})$ is defined as:
\begin{equation}
    \label{eq:CKA_sim}
    \mathbf{CKA}(\mathcal{X}, \mathcal{Y}) = \frac{\mathbb{H}(\mathcal{X}, \mathcal{Y})}{\sqrt{\mathbb{H}(\mathcal{X}, \mathcal{X})}\sqrt{\mathbb{H}(\mathcal{Y}, \mathcal{Y})}},
\end{equation}
where operator $\mathbb{H}$ stands for the unbiased Hilbert-Schmidt independence criterion proposed by \cite{song2012feature} in which $\mathbb{H}(\mathcal{X}, \mathcal{Y})$ is formulated as:
\begin{equation}
    \frac{1}{B(B-3)}(tr(\Tilde{K}\Tilde{L})+ \frac{1^T\Tilde{K}11^T\Tilde{L}1}{(B-1)(B-2)} - \frac{2}{B - 2}1^T\Tilde{K}\Tilde{L}1)
\end{equation}
where $\Tilde{K}$ and $\Tilde{L}$ are stemmed from the kernerl matrices of $\mathcal{X}$ and $\mathcal{Y}$ respectively with having $\Tilde{K}_{i,j} = (1 - \mathds{1}_{i=j})K_{ij}$ and $\Tilde{L}_{i,j} = (1 - \mathds{1}_{i=j})L_{ij}$. Noting that here the variable $B$ stands for the number of samples involved into the computation of $\mathbb{H}$, which is exactly the batch size in our application.  
}

\begin{figure}
    \centering
    \centerline{\includegraphics[width=\columnwidth]{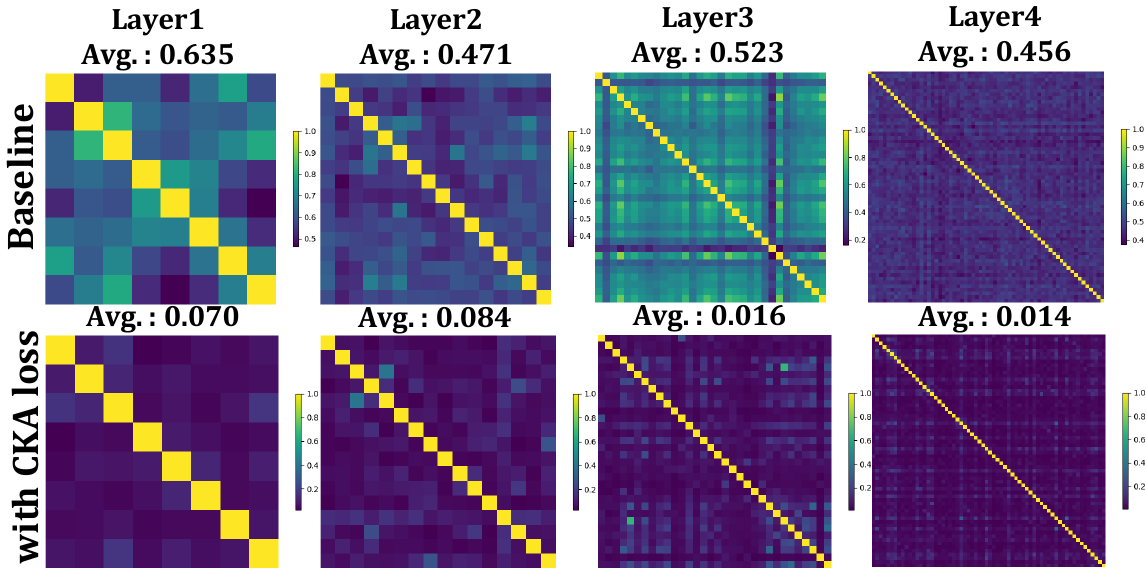}}
    \vspace{-0.3cm}
    \caption{\walon{Visualization of CKA similarities for each pair of segments from different model layers (model backbone: ResNet50; dataset: AWA2~\cite{xian2018zero}). The average for the upper triangular portion of each CKA similarity matrix is also provided accordingly. With apply our proposed CKA loss, the similarities between segments are clearly reduced, i.e. leading to more distinct concept segments.}}
    \label{fig:Base_block_sim_same_layer}
    \vspace{-0.4cm}
\end{figure}

\walon{Based on the CKA similarities between segments from $F_l$, the CKA loss $\mathcal{L}^{\mathbf{CKA}}$ of the $l$-th layer is defined as:
\begin{equation}
    \label{eq:CKA_loss}
    \mathcal{L}^{\mathbf{CKA}}(S_l) = \frac{2}{M_l(M_l-1)}\sum_{i=1}^{M_l} \sum_{j=i}^{M_l} \mathbf{CKA}(\mathit{S}_{l, i}, \mathit{S}_{l, j}),
\end{equation}
where $S_{l,i}$ and $M_l=C_l / C^\prime_l$ denotes the $i$-th concept segment and the total number of segments of $l$-th layer respectively. In \Cref{fig:Base_block_sim_same_layer} we visualize the CKA similarities between segments for different layers, highlighting the noticeable decrease in terms of CKA similarity brought by applying our proposed CKA loss.
}

\subsection{Multi-level Concept Prototype Extraction}
\label{sec:MCP_extract}
\walon{For a single input data (i.e. $B$ = 1), each concept segment $S_{l,i}$ in the $l$-th layer contains $H_l \cdot W_l$ feature vectors of length $C^\prime_l$. When it comes to the entire dataset of having $N$ data samples, there will be in total $N \cdot H_l \cdot W_l$ feature vectors for $S_{l,i}$, and every feature vector serves as an instance of the corresponding semantic behind the concept segment. In order to better outline such semantic globally behind $S_{l,i}$, we propose to identify the principal direction of all these feature vectors (i.e. analogous to the common ground among the feature vectors) by using the weighted Principal Component Analysis technique (where we weight each feature vector according to its L2-norm to account for the degree of importance).}

\walon{The resultant principal direction (i.e. the eigenvector corresponding to the largest eigenvalue of the covariance matrix built upon $N \cdot H_l \cdot W_l$ feature vectors of length $C^\prime_l$ for $S_{l,i}$) is termed as the \textbf{concept prototype}. Please note that the concept prototype is globally defined and shared within the entire dataset, while the concept segments in turn act more likely as the sample-wise or batch-wise instantiation of the corresponding semantic. Hence, we can further compute the \textbf{prototype response} of a concept segment $S_{l,i}$ (which is based on the input of a single sample or a batch of samples) with respect to the corresponding concept prototype $P_{l,i}$, following a simple procedure: Firstly the response map which records the cosine similarities of $P_{l,i}$ at each position on $S_{l,i}$ is computed, then the max-pooling is applied on the response map to obtain the prototype response. Such prototype response stands for the degree of agreement between concept segments and the concept prototype, hence signifies how likely the input sample(s) contributing to the concept segments would own the particular semantic of the target concept prototype.  
It is worth noting that, while the numerical range of original prototype response is $[-1.0, 1.0]$, we linearly map it to the range of $[0.0, 1.0]$ for better usage in the later computation.}

\walon{The extraction of concept prototypes is thoroughly applied on all the concept segments from all the model layers, leading to \textbf{multi-level concept prototypes}, in which it is one of the key factors differentiating our MCPNet from the others (where most previous works only provide single-level explanation, typically extracted from the last layer). }
\vspace{-0.5cm}
\subsection{Class-aware Concept Distribution Loss}\label{sec:CCD_loss}
\vspace{-0.2cm}
\walon{From cognitive perspective, the samples belonging to the same class ideally should have similar combination of concepts. Hence, we build upon such idea to propose the Class-aware Concept Distribution (CCD) loss, which encourages the samples of the same class to have similar distribution of concept prototypes (i.e. the distribution upon the prototype responses with respect to all the multi-level concept prototypes) while enlarging the distribution distance across different classes. In other words, such loss helps to realize the classification via leveraging our Multi-level Concept Prototype distribution (named as \textbf{MCP distribution}), while forsaking typical fully-connected-layer-based classifier with even providing better interpretability.}

\walon{Basically, with denoting the MCP distribution of an input sample/image $x_i$ as $D_i$ and the class label of $x_i$ as $y(x_i)$, we first compute the class-specific centroid MCP 
distribution $D^\mathbf{c}$ via averaging $D_i$ of all the samples $x_i$ belonging to the same class $\mathbf{c}=y(x_i)$. Then our CCD loss is defined as: 
\begin{equation}
    \begin{split}
    \mathcal{L}^{\mathbf{CCD}}(x_i) = ~&\mathbb{J}(D_i, D^{\mathbf{c}=y(x_i)}) \\
    +& \sum_{\mathbf{c'} \neq y(x_i)} \max(\mathbf{m}-\mathbb{J}(D_i, D^{\mathbf{c'}}),0),
    \end{split}
\end{equation}
where $\mathbb{J}$ stands for the Jensen-Shannon divergence (a common metric used for evaluating the distance between distributions), while $\mathbf{m}$ is the margin such that $\mathbb{J}(D_i, D^{\mathbf{c'}})$ contributes to the loss only if it is smaller than $\mathbf{m}$, and it helps avoiding a collapsed solution (which is a typical technique used the contrastive loss of metric learning).
}

\walon{The overall objective function $\mathcal{L}$ to train our model is basically the combination of both CKA and CCD losses:
\begin{equation}
    \mathcal{L} = \sum_{l=1}^{L}\mathcal{L}^{\mathbf{CKA}}(S_l) + \lambda_{CCD} \sum_{x_i \in \mathbf{X}}\mathcal{L}^{\mathbf{CCD}}(x_i),
\end{equation}
\bor{where $L$ denotes the number of layers in our model, $\mathbf{X}$ denotes the training dataset, and $\lambda_{CCD}$ denotes the weight of CCD loss.}
It is worth noting that all the concept prototypes and all the class-specific centroid MCP distributions are updated after every epoch on the training set to reflect the newest features learned by the model.
}

\subsection{Multi-Level Concept Prototypes Classifier}
\label{sec:hierarchical_structure_classifier}
\walon{As mentioned in the previous subsection, our MCPNet does not require the FC layer attached to the model end for performing classification. Instead, our MCPNet is able to classify the input sample $x_i$ simply via searching for the closest class-specific centroid MCP distribution to $D_i$:
\begin{equation}
    \Tilde{y}(x_i) = \arg\min_{\mathbf{c}}\mathbb{J}(D_i, D^\mathbf{c}).
\end{equation}}
\vspace{-1cm}
\section{Experiments}

\begin{table*}[t!]
\centering
\begin{tabular}{cccccc}
\toprule
\multirow{2}{*}{Backbone}      & \multirow{2}{*}{Methods} & \multirow{2}{*}{Explanation} & \multicolumn{3}{c}{Accuracy}          \\
                               &                          &                              & AWA2    & Caltech101 & CUB\_200\_2011 \\
\midrule
\multirow{6}{*}{ResNet50} & Baseline      & N/A          & 94.92\% & 94.21\% & 77.94\% \\
                          & ProtoTree~\cite{nauta2021neural}     & Single-Scale & 90.60\% & 72.19\% & 18.00\%\textsuperscript{$\dagger$} \\
                          & Deformable ProtoPNet~\cite{donnelly2022deformable}  & Single-Scale & 85.51\% & 93.88\% & 73.42\%\textsuperscript{$\dagger$} \\
                          & ST-ProtoPNet~\cite{wang2023learningsupport}  & Single-Scale & 93.76\% & 95.95\% & 76.34\%\textsuperscript{$\dagger$} \\
                          & PIP-Net~\cite{nauta2023pip}        & Single-Scale & 85.99\% & 87.86\% & 70.99\%\textsuperscript{$\dagger$} \\
                          & \textbf{MCPNet (Ours)}  & Multi-Scale & 93.92\% & 93.88\% & 80.15\% \\
\midrule
\multirow{6}{*}{Inception V3} & Baseline      & N/A          & 95.47\% & 96.42\% & 79.43\% \\
                             & ProtoTree~\cite{nauta2021neural}     & Single-Scale & 92.29\% & 86.02\% & 13.03\% \\
                             & Deformable ProtoPNet~\cite{donnelly2022deformable}  & Single-Scale & 92.68\% & 97.22\% & 72.99\% \\
                             & ST-ProtoPNet~\cite{wang2023learningsupport}  & Single-Scale & 93.60\% & 96.99\% & 75.25\% \\
                             & PIP-Net~\cite{nauta2023pip}        & Single-Scale & 43.82\% & 45.04\% & 6.76\% \\
                             & \textbf{MCPNet (Ours)}  & Multi-Scale & 94.62\% & 95.76\% & 78.94\% \\
\midrule
\multirow{6}{*}{ConvNeXt-tiny} & Baseline      & N/A          & 96.55\% & 96.56\% & 84.55\% \\
                               & ProtoTree~\cite{nauta2021neural}     & Single-Scale & 94.00\% & 78.82\% & 21.57\% \\
                               & Deformable ProtoPNet~\cite{donnelly2022deformable}  & Single-Scale & 91.94\% & 93.65\% & 35.05\% \\
                               & ST-ProtoPNet~\cite{wang2023learningsupport}  & Single-Scale & 94.22\% & 97.17\% & 81.84\% \\
                               & PIP-Net~\cite{nauta2023pip}        & Single-Scale & 93.80\% & 96.61\% & 82.74\% \\
                               & \textbf{MCPNet (Ours)}  & Multi-Scale & 95.61\% & 95.95\% & 83.45\% \\
\bottomrule
\end{tabular}
\vspace{-0.3cm}
\caption{The classification performance evaluation on AWA2, Caltech101, and CUB\_200\_2011 benchmarks with different backbone choices. The baseline represents the typical classification without any explanation capability.  
\bor{\textsuperscript{$\dagger$}The discrepancies with respect to the accuracies reported in their original papers are caused by using different pretrained weights (here we adopt the pertaining on ImageNet). }
}
\label{tab:MCP_performance}
\vspace{-0.5cm}
\end{table*}
\noindent\textbf{Datasets.}
Three datasets are adopted for our evaluation:
\begin{itemize}
    \item \textbf{AWA2} \cite{xian2018zero} is dataset which was originally proposed for evaluating the zero-shot classification task. It consists of 37322 images with 50 categories, each is additionally annotated with 85 attributes. We split a quarter of the entire dataset as the test set, while the rest is taken as the training set. Please note that the attribute labels are not used in our model training.
        
    \item \textbf{Caltech101} \cite{li2022caltech} is a classification dataset that has 9146 images from 101 distinct categories, in which each class has roughly 40 to 800 images. We randomly draw a quarter of images of each class to form the test set, while the rest becomes the training set.
    
    \item \textbf{CUB\_200\_2011} \cite{wah2011cub} is a fine-grained image classification dataset that contains 11788 images of 200 bird classes/species (and additional has 312 binary labels of attributes), where 5,994 images are used for training while the other 5,794 images are for testing. Please note that, we only use the training images and corresponding class/species labels for performing our model training without using the additional attribute annotations.

\end{itemize}
\vspace{-0.2cm}
\subsection{Implementation Details}
\paragraph{Layer Selections.} We adopt three off-the-shelf models, ResNet50 \cite{he2015delving}, and Inception v3 \cite{szegedy2016rethinking}, ConvNeXt-tiny \cite{liu2022convnet} to do the experiments. The layers selected for each model are shown in the supplementary.
\vspace{-0.3cm}
\paragraph{Training hyper-parameters.}
\bor{We conduct training for ResNet50 and InceptionNet V3 over 100 epochs, using Adam optimization with a weight decay set to 1e-4. The learning rate follows a decay rate of 10 every 40 epochs, beginning at 1e-4.} For ConvNeXt-tiny, we utilized the official training code with 100 epochs. The dimension for concept segments, denoted as $C^\prime$, is set to 32 for ResNet50 and InceptionNet v3 and 16 for ConvNeXt-tiny. The Class-aware Concept Distribution (CCD) loss margin is set to 0.01 for ResNet50 and InceptionNet v3 and 0.05 for ConvNeXt-tiny. 
\bor{As the original numerical range of CCD loss is way smaller than that of CKA loss, we set $\lambda_{CCD}$ to 100 to balance between these two loss functions.}

\begin{figure*}[]
    \centering
    \includegraphics[width=0.99\textwidth]{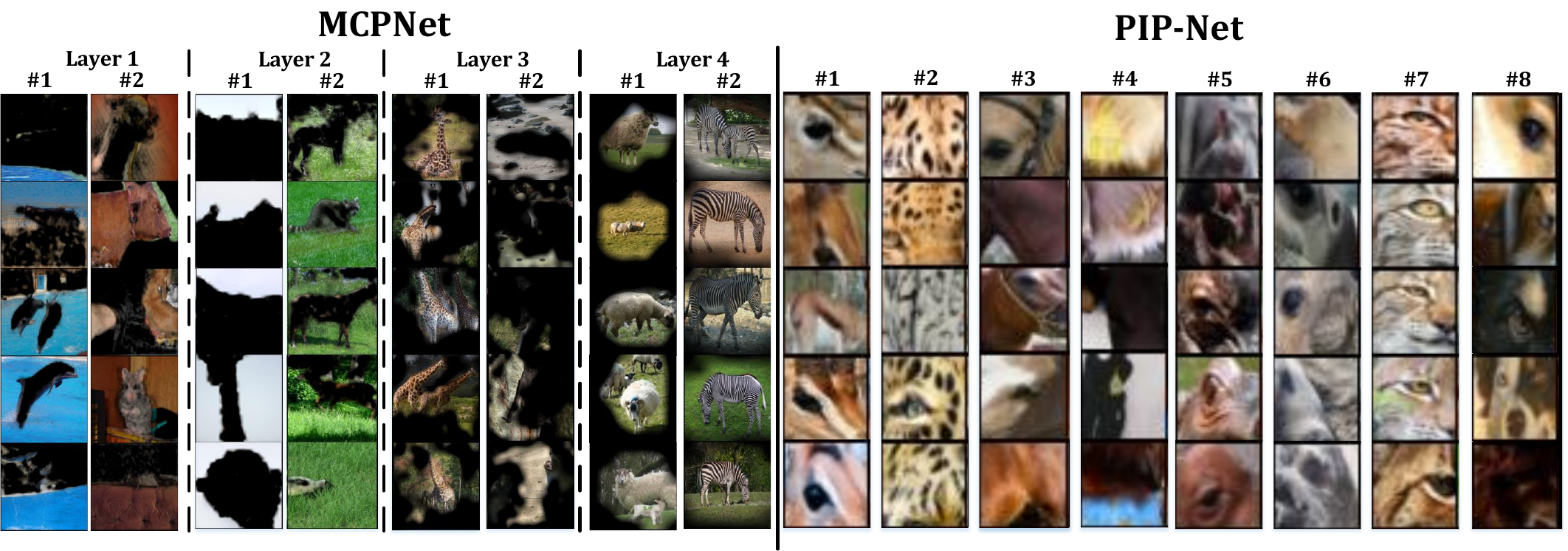}
    \vspace{-0.3cm}
    \caption{\bor{Concept prototype examples from MCPNet and PIP-Net \cite{nauta2023pip}. We show the top-5 responses for the sampled concept prototypes. For MCPNet, the concept prototype from various layers generates explanations in different scales, e.g. color-like explanations in low-layer and object-like explanations in high-layer. On the contrary, PIP-Net \cite{nauta2023pip} only provides single-scale patch-level explanations.}}
    \label{fig:Sample_nodes}
    \vspace{-0.4cm}
\end{figure*}

\subsection{Quantitative Results}
\bor{This section explains how MCPNet provides explanations at multiple levels without compromising performance, comparing to the typical training paradigm coupled with a fully connected (FC) classifier. Basically, the conventional approach trains the model using cross-entropy loss, focusing the model's learning on features in the final layer for image differentiation -- a technique widely used in prior methods. In contrast, MCPNet classifies images based on the distribution formed by the response of multi-level concept prototypes, considering concepts of various scales.}

\bor{In \Cref{tab:MCP_performance}, we present the primary quantitative outcomes across three classification datasets, comparing MCPNet with the baseline and other methods in the ProtoPNet series. MCPNet matches or surpasses their performance by categorizing images through alignment with the nearest class-specific MCP distribution and delivers explanations across multiple levels. Conversely, alternative methods offer explanations at a single scale, primarily concentrating on the model's final layer to produce object-centric explanations. The findings further demonstrate MCPNet's versatility, as it can be integrated into diverse models to yield robust performance and multi-level explanations.}

\begin{table}[]
\centering
\begin{tabular}{ccc}
\toprule
Dataset & Method & Accuracy \\ 
\midrule
 \multirow{6}{*}{AWA2} & Baseline & 60.55\% \\ 
 & ProtoTree \cite{nauta2021neural} & 33.68\% \\ 
 & Deformable ProtoPNet \cite{donnelly2022deformable} & 19.71\% \\ 
 & ST-ProtoPNet \cite{wang2023learningsupport} & 30.15\% \\ 
 & PIP-Net \cite{nauta2023pip} & 26.17\% \\ 
 & \textbf{MCPNet (Ours)} & \textbf{73.79}\% \\ 
\bottomrule

\end{tabular}
\vspace{-0.3cm}
\caption{The 5-shot classification performance with ResNet50 backbone on AWA2 datasets. The baseline represents the typical classification without any explanation capability.}
\label{tab:Few_shot}
\vspace{-0.5cm}
\end{table}
\vspace{-0.5cm}
\paragraph{5-shot Classification.} 
In the 5-shot experiments, the datasets are divided into seen and unseen sets. The model is trained on the seen set, from which we derive global concept prototypes. Within the unseen set, 5 images from each category are randomly selected to create the class-specific MCP distributions for classes in the unseen set. The remaining images are utilized to assess the model's efficacy in 5-shot classification.

We evaluate the accuracy of our approach against the baseline and methods from the ProtoPNet series. For the baseline, we apply the same classification mechanism used in MCPNet to images in the unseen set. For the ProtoPNet series methods, the model was trained on the seen set using default settings, followed by a single epoch of training on the images which are arranged similarly to those used for creating the MCPNet's class MCP distribution. This one-epoch training is chosen to align with MCPNet for its processing the images in a single pass.

In \Cref{tab:Few_shot}, we demonstrate our method achieves competitive outcomes in the 5-shot experiments. This indicates that the concepts we extract are sufficiently generalized to discern between classes, including those not involved in the concept extraction process.

\subsection{Qualitative Results.}
\bor{In this section, we introduce how we recognize the meaning of concept prototypes and how MCPNet explains the images or even the classes with the MCP distributions.}
\vspace{-0.5cm}
\paragraph{Concept Prototype meanings.} 
\bor{To intuitively grasp each concept prototype's essence, we visually depict them through images that elicit the highest top-5 responses for a given concept within the dataset. By overlaying these images with the response map described in \Cref{sec:MCP_extract}, the visible regions help us identify the features highlighted by the concept prototype. Our findings are contrasted with those from PIP-Net~\cite{nauta2023pip} in \Cref{fig:Sample_nodes} -- unlike PIP-Net~\cite{nauta2023pip} where the explanations are typically tied to specific object parts, MCPNet offers explanations on multiple levels derived from various layers of the model. These explanations range from high-level concepts, such as parts of an object, to low-level attributes, such as color. Additional examples are provided in \Cref{fig:Sample_nodes2} and in the supplementary material.}

\begin{figure*}[]
    \centering
    \includegraphics[width=0.99\textwidth]{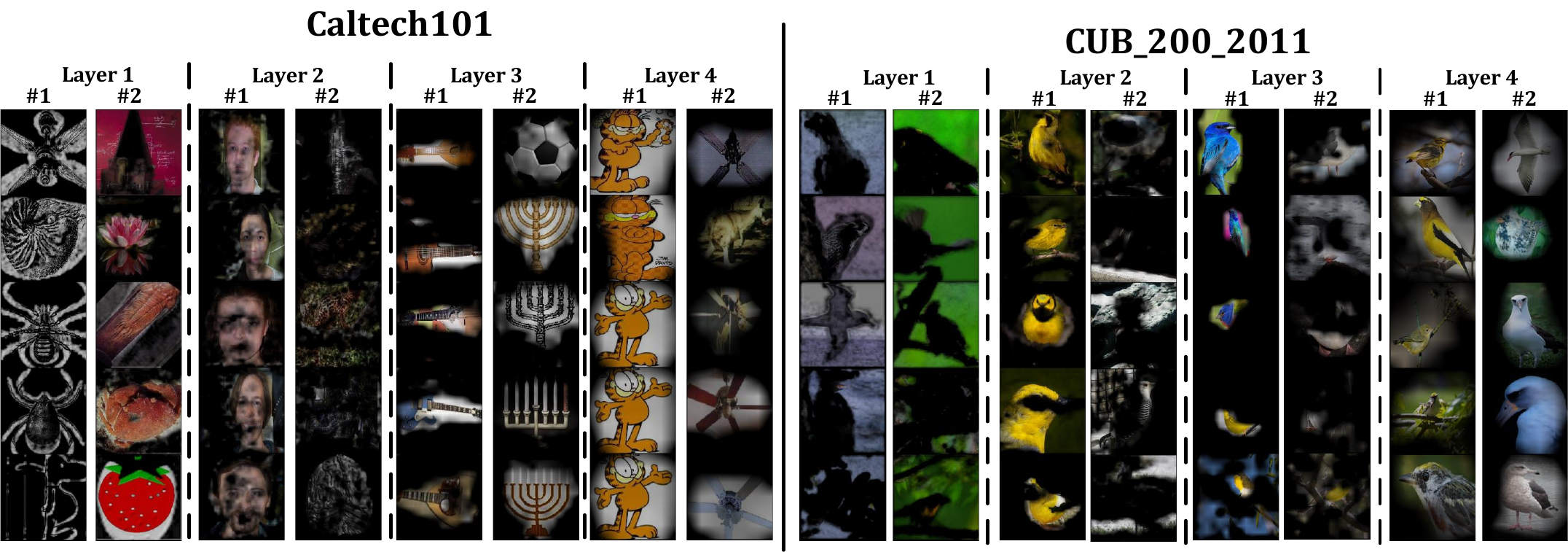}
    \vspace{-0.4cm}
    \caption{The sampled multi-level concept prototypes learnt by our proposed MCPNet. (Backbone : ResNet50)}
    \label{fig:Sample_nodes2}
    \vspace{-0.3cm}
\end{figure*}

\paragraph{MCP distribution explanations.} 
\bor{Here we provide explanation upon how our MCP distribution is used to interpret the model's classification of images. For each image, we obtain the corresponding MCP distribution by evaluating the interactions between the extracted concept prototype and its related concept segment, which shows the concepts contained in the image. As depicted in Figure \ref{fig:teaser_figure}, we determine the image's classification by comparing its distribution to the class-specific MCP distribution. This process involves identifying concept responses that show a high degree of similarity to a particular category, suggesting that the image's concepts resemble those of the target category. This forms the basis of our explanation for why an image is categorized into a specific class. 
For each class, the class-specific MCP distribution -- which averages the MCP distributions of images within that class -- highlights prevalent concept responses, with higher responses denoting concepts frequently associated with that class.}

\begin{figure}[h]
    \centering
    \includegraphics[width=0.45\textwidth]{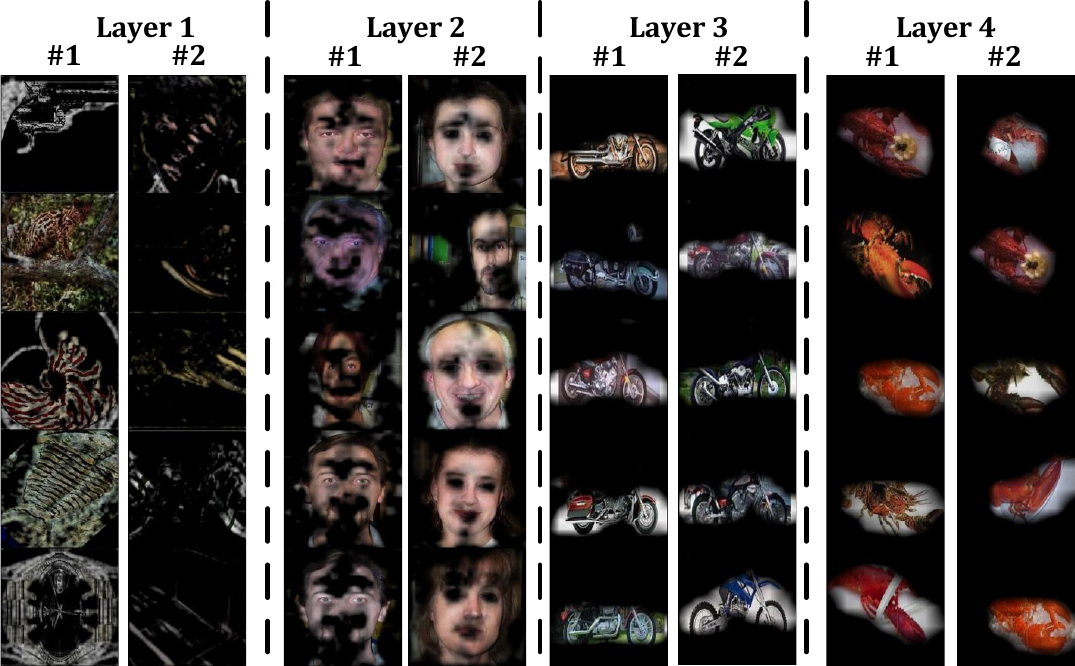}
    \vspace{-0.3cm}
    \caption{Showing the multi-level concept prototypes learnt by MCPNet variant with only the CCD loss, in which there are more duplicate prototypes than the ones learnt by the full MCPNet (i.e. the CKA loss is also adopted).}
    \label{fig:TC_sample}
\end{figure}

\subsection{Ablation Study}
\label{sec:ablation_study}
\begin{table}[]
\centering
\begin{tabular}{ccc}
\toprule
CKA loss & CCD loss & Accuracy \\  
\midrule
 \cmark &   & 44.85\% \\  
   & \cmark & 94.21\%  \\ 
 \cmark & \cmark & 93.88\%  \\
\bottomrule
\end{tabular}
\vspace{-0.3cm}
\caption{Ablation study on the effect of CKA loss and CCD loss. It shows the MCPNet classification accuracy on Caltech101 with ResNet50 backbone.}
\label{tab:ablation}
\vspace{-0.5cm}
\end{table}
\paragraph{The effect of the different losses.}  
\vspace{-0.1cm}
\bor{\Cref{tab:ablation} shows the performance impacts of including or excluding each proposed loss. While the CKA loss is dedicated to disentangling concept segments, its singular use results in performance that falls below the baseline, as it lacks the capability to classify images effectively. The integration of the proposed CCD loss, on the other hand, leads to an improvement in accuracy over the baseline. However, without the CKA loss, there is an observed increase in similarity among concept segments, leading to duplicated concept prototypes, a phenomenon depicted in \Cref{fig:TC_sample}. The combined application of CCD and CKA losses outperforms the exclusive use of CCD loss, achieving more distinct concept prototypes with only a slight compromise in performance.}

\paragraph{The effect of channel size.} 
\vspace{-0.4cm}
\bor{\Cref{tab:Diff_cha} presents the impact of different channel sizes on the performance of our model, with a comparison among 32, 16, and 8 channels. The model with 32 channels offers half the concept segments of the 16-channel setup and a quarter of those in the 8-channel configuration. On coarse-grained datasets like AWA2 and Caltech101, the performances of models with 32 and 8 channels are comparable, indicating that the number of concept segments with 32 channels suffice for capturing class distinctions in these scenarios. In contrast, the 8-channel setup shows enhanced performance on the fine-grained CUB dataset, likely attributable to the higher number of concept segments that better capture the nuanced differences within the dataset.}

\begin{table}[]
\centering
\begin{tabular}{ccc}
\toprule
Dataset & Channel & Accuracy \\  
\midrule
 \multirow{3}{*}{AWA2} & 32 & 93.92\% \\ 
 & 16 & 93.95\% \\ 
 & 8 & 93.58\% \\ 
 \midrule
 \multirow{3}{*}{Caltech101} & 32 & 93.88\% \\ 
 & 16 & 93.79\% \\ 
 & 8 & 93.51\% \\ 
 \midrule
 \multirow{3}{*}{CUB\_200\_2011}  & 32 & 80.15\% \\ 
 & 16 & 80.19\% \\ 
 & 8 & 81.22\% \\ 
\bottomrule
\end{tabular}
\vspace{-0.3cm}
\caption{The ablation study with different channel sizes of the concept segments (Backbone: ResNet50).}
\label{tab:Diff_cha}
\vspace{-0.5cm}
\end{table}

\section{Conclusion}
\vspace{-0.2cm}
\noindent In this paper, we propose Multi-Level Concept Prototype Classifier (MCPNet), an inherently interpretable method which learns multi-layer concept prototypes without reliance on predefined concept labels. In addition to having more comprehensive multi-level model explanations, our MCPNet is experimentally shown to provide a classification paradigm which is able to achieve comparable performance as the typical fully-connected-layer-based classifier while achieving better generalizability upon unseen classes. 

\maketitlesupplementary
\setcounter{page}{1}

\section{Layer selections}
\begin{table}[h]
\centering
\begin{tabular}{c|c|c|c}
Layer & ResNet50 & Inception v3 & ConvNeXt tiny \\ \hline
1 & layer1 & Conv2d\_4a\_3x3 & stages[0] \\
2 & layer2 & Mixed\_6a & stages[1] \\
3 & layer3 & Mixed\_7a & stages[2] \\
4 & layer4 & Mixed\_7c & stages[3]
\end{tabular}
\caption{Selected layers for the three backbone models used in this paper.}
\label{tab:sel_layer}
\end{table}
In \Cref{tab:sel_layer}, we present the selected layers for ResNet50, InceptionV3, and ConvNeXt-tiny, with the layer names following the PyTorch code. In the case of ResNet50 and ConvNeXt, the selected layers are based on the original architecture design, where each layer corresponds to one convolutional block. For InceptionV3, we partition the model into four parts, treating each as a distinct layer.
\section{Training process}
Our complete training workflow is depicted in \Cref{alg:train}. We initiate the process by extracting the multi-level concept prototypes (MCP) and class-specific multi-level concept prototype distributions (MCP distribution), where the backbone model is initialized by using the weights pre-trained on ImageNet. 
The term "CalMCP" in Lines 2 and 11 of \Cref{alg:train} denotes the process of computing the MCP using weighted Principal Component Analysis (PCA) to capture the global semantic information from the segments, as detailed in \Cref{sec:MCP_extract}.
The term "CalMCPDist" in Lines 3 and 12 of \Cref{alg:train} represents the procedure of averaging the MCP distribution across images within the same class, yielding the class-specific MCP distribution that encapsulates the predominant concept distribution, 
as explained in \Cref{sec:CCD_loss}.
Throughout the training process, these extracted MCP and MCP distributions are employed to calculate the loss and we update them every epoch to incorporate the latest features learned by the model.

\SetKwInput{KwInput}{Input}                
\SetKwInput{KwOutput}{Output}              

\begin{algorithm}
\DontPrintSemicolon
\KwInput{Training set $\mathcal{T} = \{\mathcal{T}_1, \mathcal{T}_2, ..., \mathcal{T}_n\}, (x_1^\mathcal{T}, y_1^\mathcal{T}) = \mathcal{T}_1$, validation set $\mathcal{V} = \{\mathcal{V}_1, \mathcal{V}_2, ..., \mathcal{V}_n\}, (x_1^\mathcal{V}, y_1^\mathcal{V}) = \mathcal{V}_1$, Epochs $E$, Concept segment size $C^\prime$.
}

Initial model $f$ weight $w_f$ with pre-trained on ImageNet and remove the fully connected layer.

\tcc{Compute the multi-level concept prototypes (MCP). Mentioned in \Cref{sec:MCP_extract}.}
$\mathcal{MCP}$ = CalMCP($f$, $\mathcal{T}$);

\tcc{Compute the class-specific MCP distribution $\mathcal{D} = \{\mathcal{D}^{1}, \mathcal{D}^{2}, ..., \mathcal{D}^{y_n}\}$. Mentioned in \Cref{sec:CCD_loss}.}
$\mathcal{D}$ = CalMCPDist($\mathcal{T}$, $\mathcal{MCP}$);

\For{epoch $\in$ \{1,..., $E$\}}{

    \tcc{Train phase}
    \For {$(x_b, y_b) \in \{\mathcal{T}_1, ...,\mathcal{T}_n\}$}{
        $F_1, F_2, F_3, F_4 = f(x_b)$ 
        
        split segments $S_l \leftarrow F_l, l \in \{1, 2, 3, 4\}$

        compute loss $\mathcal{L}^{\textbf{CKA}}(S_l)$

        compute loss $\mathcal{L}^{\textbf{CCD}}(x_b)$ with class-specific MCP distribution $\mathcal{D}$

        Minimize losses by updating $w_f$
    }

    \tcc{Re-calculate MCP}
    $\mathcal{MCP}$ = CalMCP($f$, $\mathcal{T}$)
    
    \tcc{Re-calculate class-specific MCP distribution}
    $\mathcal{D}$ = CalMCPDist($\mathcal{T}$, $\mathcal{MCP}$)

        
}

\caption{Training a MCPNet}
\label{alg:train}
\end{algorithm}

\begin{figure}
    \centering
    \includegraphics[width=0.5\textwidth]{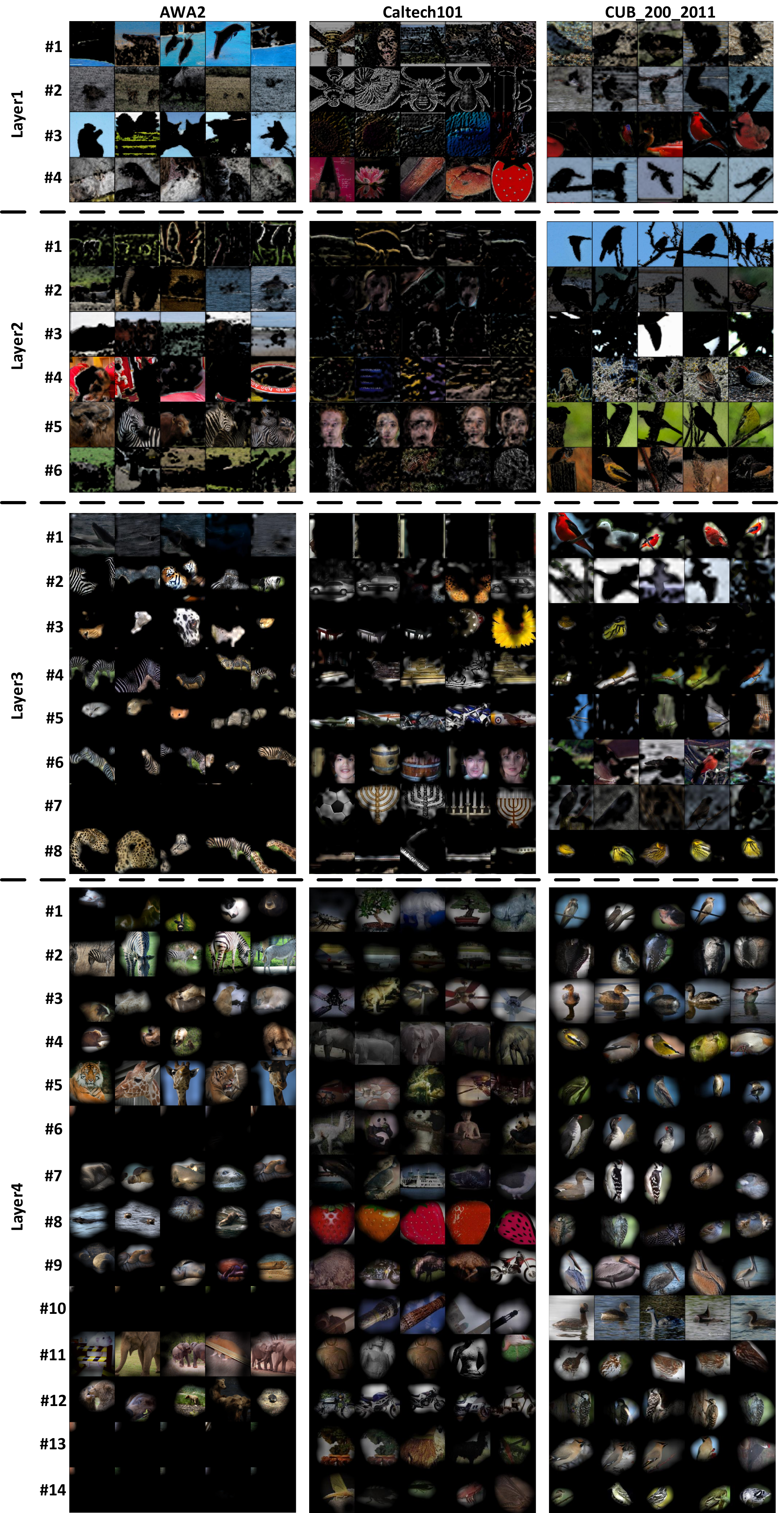}
    \caption{Multi-level concept prototype samples on three datasets. (Backbone: ResNet50)}
    \vspace{-0.5cm}
    \label{fig:resnet50_sample}
\end{figure}

\begin{figure}
    \centering
    \includegraphics[width=0.5\textwidth]{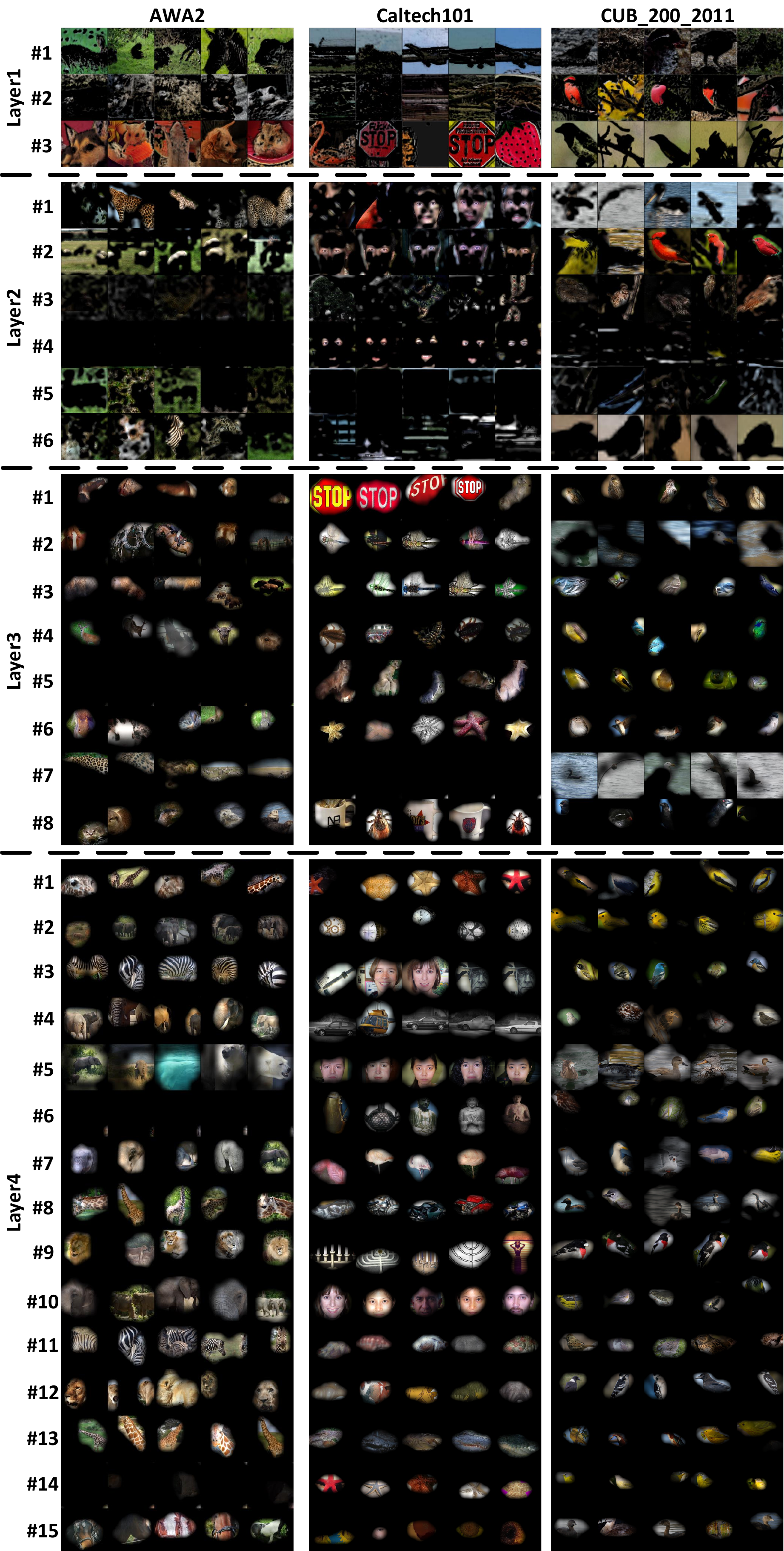}
    \caption{Multi-level concept prototype samples on three datasets. (Backbone: Inception V3)}
    \vspace{-0.5cm}
    \label{fig:inceptionv3_sample}
\end{figure}

\begin{figure}
    \centering
    \includegraphics[width=0.5\textwidth]{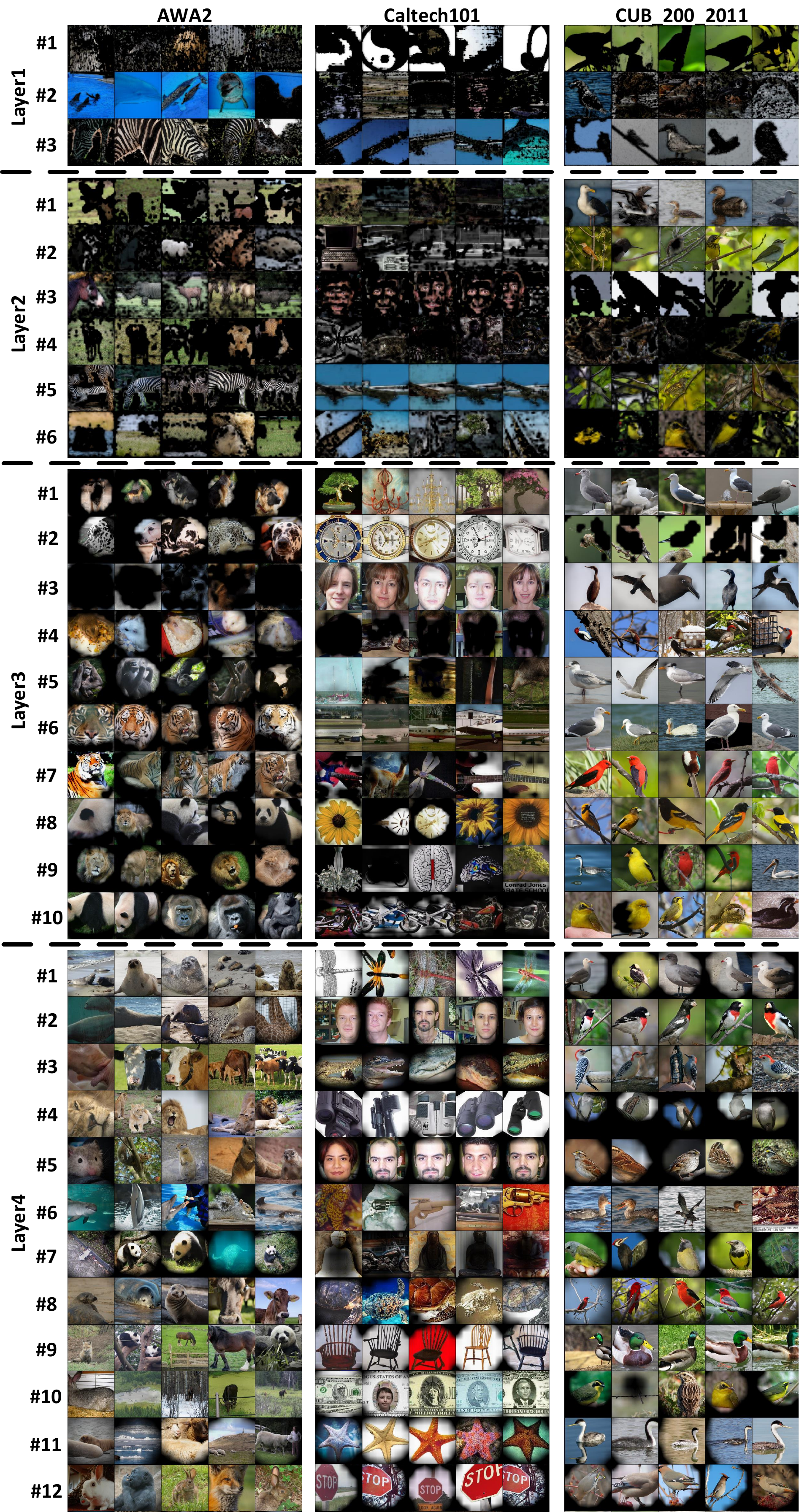}
    \caption{Multi-level concept prototype samples on three datasets. (Backbone: ConvNeXt-tiny)}
    \vspace{-0.5cm}
    \label{fig:convnext_sample}
\end{figure}

\begin{figure}
    \centering
    \includegraphics[width=0.4\textwidth]{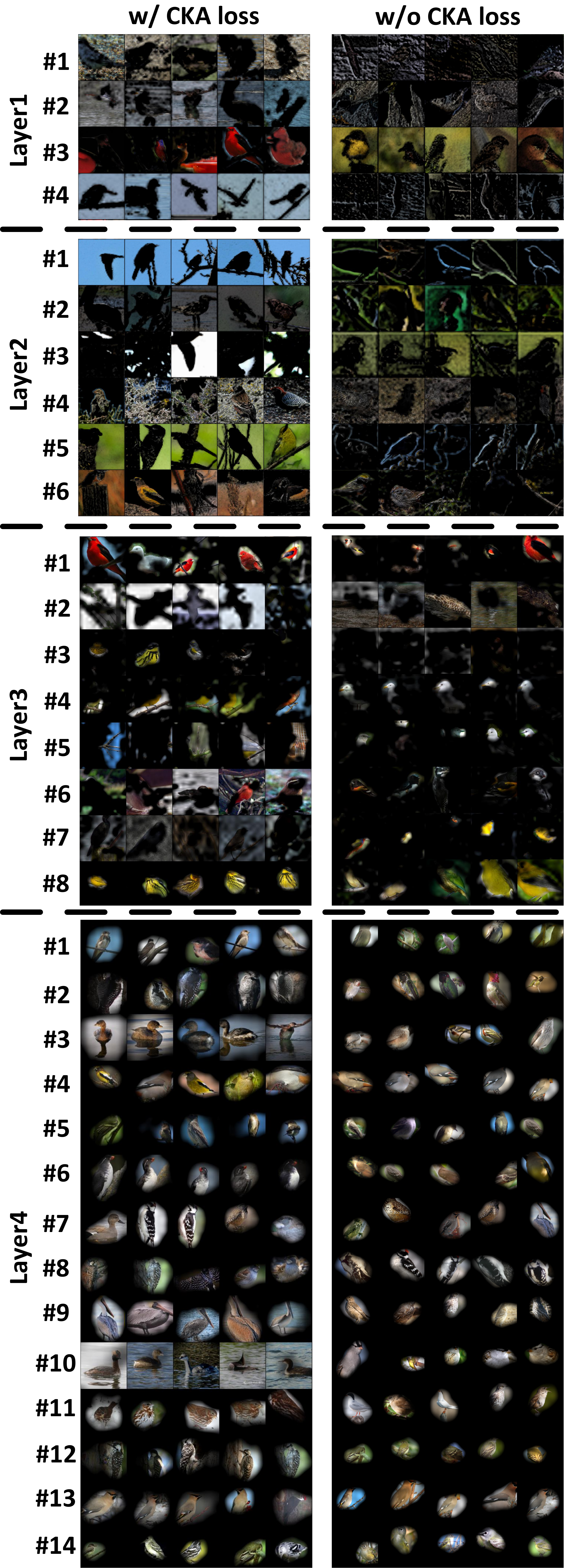}
    \caption{More samples of w/ and w/o CKA loss(Backbone: ResNet50; Dataset: CUB\_200\_2011).}
    \vspace{-0.5cm}
    \label{fig:ablation_sample}
\end{figure}

\section{More benchmarks}
In \Cref{tab:benchmark}, we present additional performance comparisons between our Multi-Level Concept Prototypes Classifier (MCPNet) and the ProtoPNet series methods \cite{nauta2021neural, donnelly2022deformable, wang2023learningsupport, nauta2023pip} across three datasets with different backbone architectures. We retrained all models using their source code (i.e. their models are also pretrained on ImageNet dataset with batch size set to 64, following the same setting as ours)
and full images (i.e. no cropping) for fair comparisons.
The results indicate that MCPNet achieves comparable performance in various scenarios by offering multi-level explanations, thereby providing a more comprehensive understanding of the deep-learning black box.

Please kindly note that, the results of our retraining ProtoPNet series methods on ResNet50 differ from the reported results in their papers since their original experimental settings adopt the pretraining on iNaturalist 2017~\cite{van2018inaturalist} with some of them utilizing cropped images or different batch size during training.
The notably low performance of PIP-Net~\cite{nauta2023pip} on ResNet34 may be attributed to their method not being optimized for a smaller model (while the training of ConvNeXt based on our experimental setting does achieve the similar result as in the original PIP-Net~\cite{nauta2023pip} paper, hence verifying the correctness of our implementation). 
Regarding ProtoTree~\cite{nauta2021neural} on CUB\_200\_2011, the performance appears to be particularly sensitive to pretraining on iNaturalist 2017, as we could not achieve satisfactory results with pretraining on ImageNet alone. It is worth noting that the original paper of ProtoTree~\cite{nauta2021neural} only provides quantitative results for ResNet50. 

\bor{
We present the performance comparison with the ProtoPNet series methods on the CUB\_200\_2011 dataset in \Cref{tab:MCP_performance_large}, where ResNet50 is adopted as backbone while being now pretrained on the iNaturalist 2017 dataset (i.e. in comparison to the results in \Cref{tab:benchmark} where the pretraining is based on ImageNet dataset, here we follow the common setting in most of ProtoPNet series methods to have ResNet50 pretrained on iNaturalist 2017 dataset). We can observe that, even with different pretrained weights for the ResNet50 backbone, our proposed MCPNet manages to maintain performance that is on par with other methods.}

\begin{figure*}
    \centering
    \includegraphics[width=1.0\textwidth]{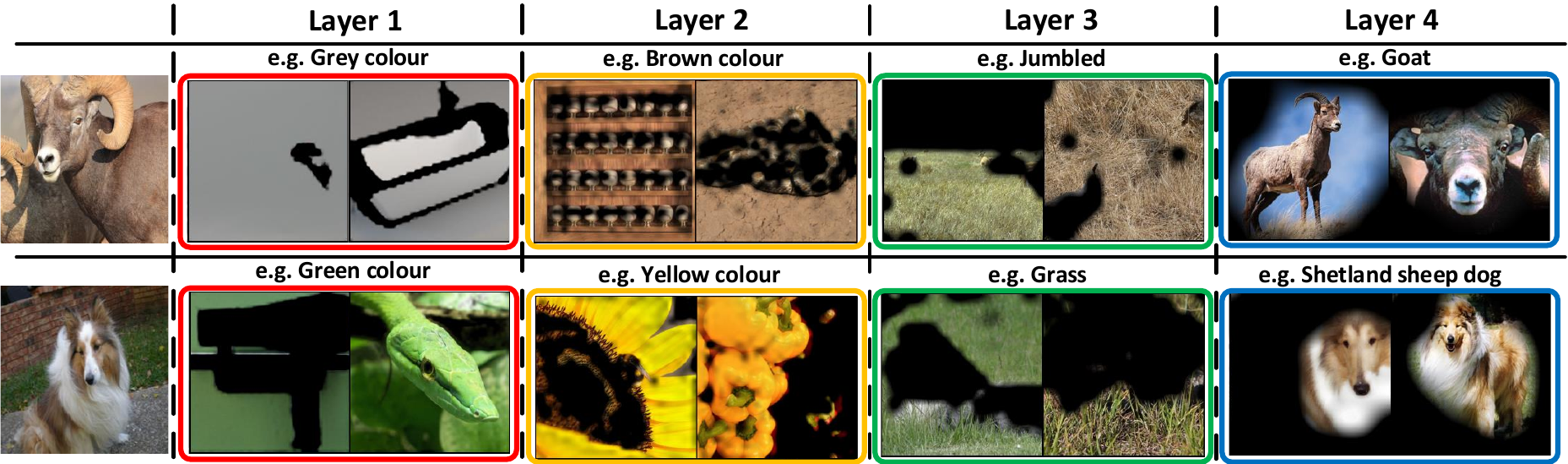}
    \caption{\bor{The interpretation of images (leftmost column) based on the multi-scale prototypes (extracted on different layers of the classification model) learnt from the large-scale sampled ImageNet-1K dataset by our MCPNet (using ResNet50 as the backbone), e.g. identifying colors such as white and green from the lower-level layers, and recognizing entities like goats and Shetland sheepdogs from the higher-level layers.}}
    \label{fig:rebuttal_ImageNet_sample12}
    \vspace{-0.3cm}
\end{figure*}

\vspace{-0.5cm}
\paragraph{Large-scale dataset}
\bor{We also evaluate our model on the large-scale dataset, denoated as \textbf{sampled ImageNet-1K}, which is built by randomly sampling one-tenth of the images for each category in the ImageNet training set while keeping the full ImageNet evaluation set. As shown by the results summarized in \Cref{tab:MCP_performance_large}, our method is capable of learning discriminative MCP distributions and meaningful MCPs on such large-scale dataset. In \Cref{fig:rebuttal_ImageNet_sample12}, we also provide example visualizations of multi-scale prototypes learnt by MCPNet on the large-scale sampled ImageNet-1K dataset.
It is worth noting that, while most previous prototype-based methods focus on learning class-specific prototypes with often having their inter-class prototypes enforced to be orthogonal, such orthogonality becomes quite challenging to maintain when the number of classes increases thus typically leading to diminished performance.
}




\begin{table}[t!]
\centering
\scalebox{0.9}{
\begin{tabular}{lccc}
\toprule
    Method & \begin{tabular}[c]{@{}c@{}}CUB\\ (iNaturalist)\end{tabular} & \begin{tabular}[c]{@{}c@{}}Sampled \\ ImageNet-1K\\ (None)\end{tabular} \\
    \midrule
    ProtoTree [\textcolor{green}{13}] & 77.22\% & 9.07\% \\
    Deformable ProtoPNet [\textcolor{green}{2}] & 85.66\% & 1.48\%\\
    ST-PrototPNet [\textcolor{green}{28}] & 87.63\% & 58.15\% \\
    PIP-Net [\textcolor{green}{14}] & 82.33\% & 0.10\% \\
    \textbf{MCPNet (Ours)} &  86.28\% & 61.73\% \\
    \bottomrule    
\end{tabular}
}
\vspace{-0.3cm}
\caption{\bor{We here provide two experimental results: 1) The middle column reveals how MCPNet (as well as the ProtoPNet series methods), while the backbone is pretrained on the iNaturalist dataset, performs on the CUB dataset; 2) The rightmost column details the performance when various models are trained on a sampled subset of ImageNet-1K and then evaluated on the entire ImageNet-1K validation set. (Backbone : ResNet50).}}
\label{tab:MCP_performance_large}
\vspace{-0.3cm}
\end{table}
\begin{table*}[]
\centering
\begin{tabular}{cccccc}
\hline
\multirow{2}{*}{Backbone} & \multirow{2}{*}{Methods} & \multirow{2}{*}{Explanation} & \multicolumn{3}{c}{Accuracy} \\
 &  &  & AWA2 & Caltech101 & CUB\_200\_2011 \\ \hline
\multirow{6}{*}{ResNet34} & Baseline      & N/A          & 94.60\% &  94.78\% & 77.89\% \\
                          & ProtoTree~\cite{nauta2021neural}     & Single-Scale & 90.33\% & 77.71\% & 15.79\% \\
                          & Deformable ProtoPNet~\cite{donnelly2022deformable}  & Single-Scale & 90.55\% & 95.95\% & 74.53\% \\
                          & ST-ProtoPNet~\cite{wang2023learningsupport}  & Single-Scale & 93.56\% & 96.24\% & 77.65\% \\
                          & PIP-Net~\cite{nauta2023pip}        & Single-Scale & 8.00\% & 44.52\% & 7.65\% \\
                          & \textbf{MCPNet (Ours)}  & Multi-Scale & 93.02\% & 93.32\% & 76.98\% \\ \hline

\multirow{6}{*}{ResNet152} & Baseline      & N/A          & 95.76\% & 95.76\% & 78.82\% \\
                           & ProtoTree~\cite{nauta2021neural}     & Single-Scale & 92.48\% & 87.72\% & 20.88\% \\
                           & Deformable ProtoPNet~\cite{donnelly2022deformable}  & Single-Scale & 91.16\% & 93.65\% & 76.27\% \\
                           & ST-ProtoPNet~\cite{wang2023learningsupport}  & Single-Scale & 93.95\% & 96.09\% & 79.01\% \\
                           & PIP-Net~\cite{nauta2023pip}        & Single-Scale & 64.44\% & 69.88\% & 26.87\% \\
                           & \textbf{MCPNet (Ours)}  & Multi-Scale & 94.28\% & 94.54\% & 80.79\% \\ \hline

\end{tabular}
\caption{The comparison between our MCPNet and various methods on three benchmark datasets: AWA2, Caltech101, CUB\_200\_2011.}
\vspace{-0.3cm}
\label{tab:benchmark}
\end{table*}
\vspace{-0.3cm}
\section{Prototype Visualizations}
To grasp the meaningful concepts represented by the concept prototypes in MCPNet, we visualize each prototype with the top-5 response images from the dataset, as illustrated in Figure \ref{fig:resnet50_sample}, \Cref{fig:inceptionv3_sample}, and  \Cref{fig:convnext_sample} (respectively for the backbones of ResNet50, Inception V3, and ConvNeXt-tiny). Each row of images corresponds to a single concept prototype, showcasing prototypes from multiple levels that represent explanations of different scales for the model.

It is noteworthy that \textit{null} concept prototypes exist in some experiments, as exemplified by the $\#7$ prototype in layer 3 and the $\#6, \#10, \#13, \#14$ prototypes in layer 4 from AWA2, which are shown in \Cref{fig:resnet50_sample}. 
The presence of \textit{null} concepts may suggest that the model has learned a sufficient number of concept prototypes, which is fewer than our predefined number of concepts in that layer, to effectively perform the image classification. Moreover, as the dataset complexity increases, concept prototypes are encouraged to learn more discriminative concepts for accurate image classification thus leading to the decrease for the number of \textit{null} prototypes, as observed in the concept prototypes from Caltech101 and CUB\_200\_2011 in \Cref{fig:resnet50_sample}.

\begin{figure*}
    \centering
    \includegraphics[width=1.0\textwidth]{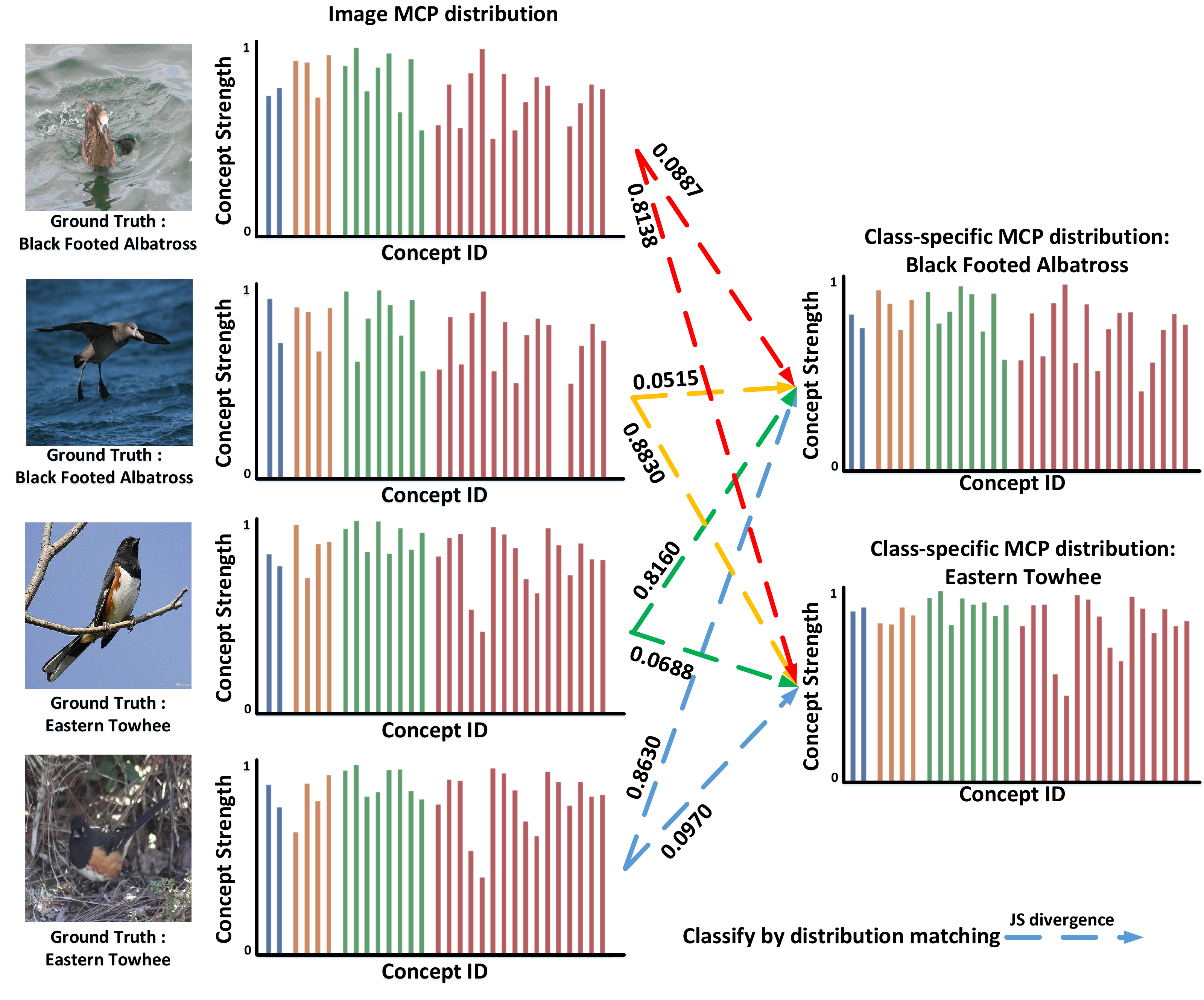}
    \caption{The demonstration of the image classification process in MCPNet, which involves matching image MCP distributions to the class-specific MCP distribution.}
    \label{fig:instance_samples}
\end{figure*}

\begin{figure*}
    \centering
    \includegraphics[width=1.0\textwidth]{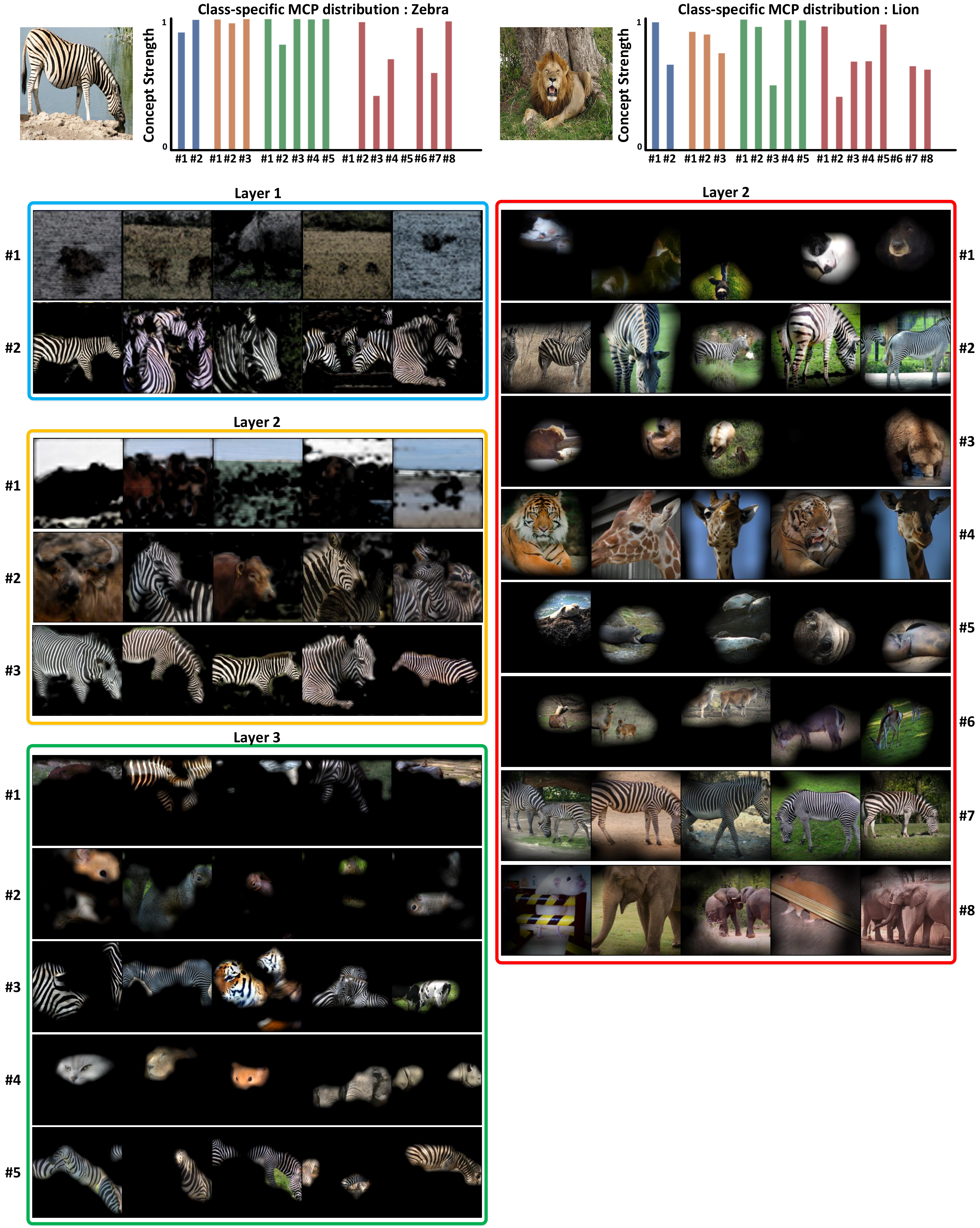}
    \caption{Image-wise explanation. For each image, the explanation is derived by calculating the concept strength to form the MCP distribution. The concept strength indicates whether the concept prototypes are prevalent in the image.}
    \label{fig:coarse_samples}
\end{figure*}

\begin{figure*}
    \centering
    \includegraphics[width=1.0\textwidth]{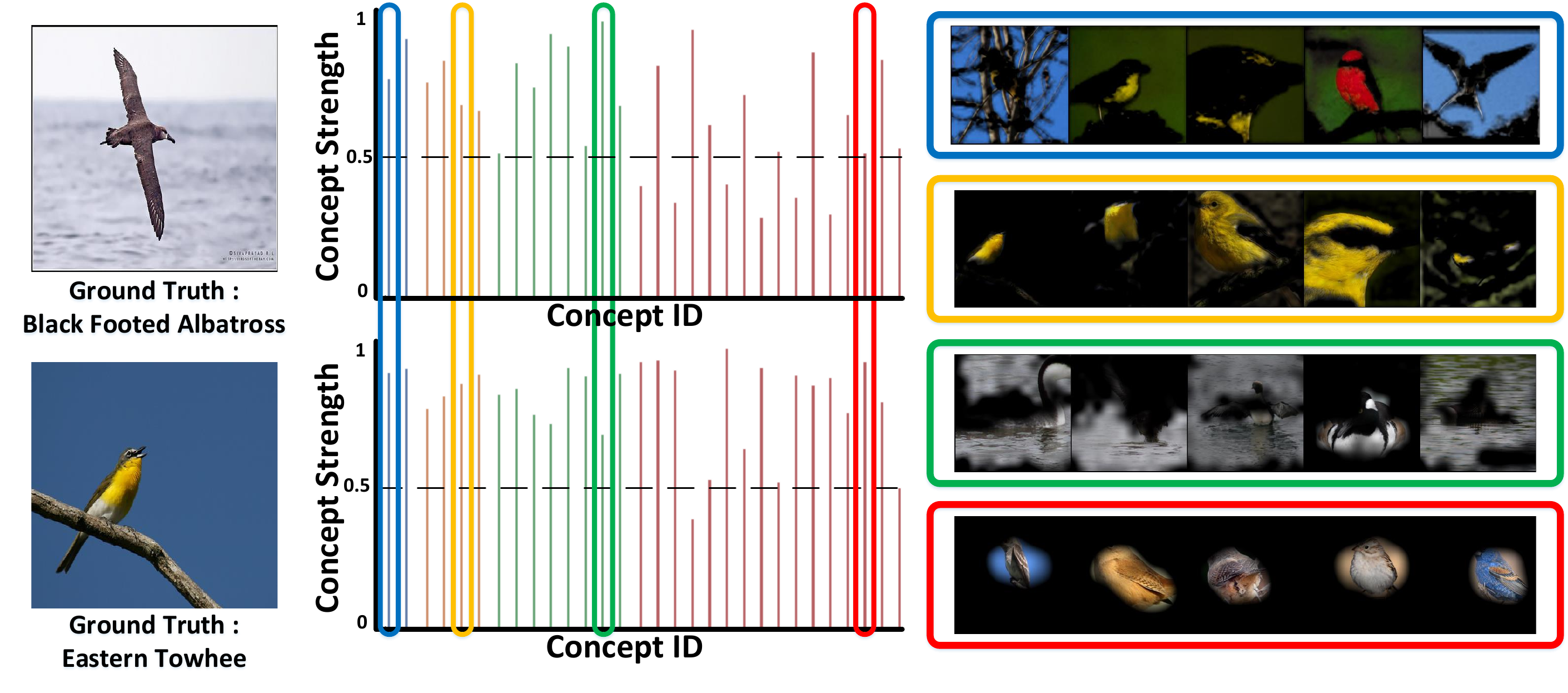}
    \caption{Illustrates the disparity between class-specific MCP distributions on CUB\_200\_2011 (backbone: ResNet50). We select concepts that elicit significantly different responses between the two class-specific MCP distributions to showcase the class-wise explanation based on the presence or absence of concept prototypes. The colors (blue, yellow, green, and red) represent layers 1 to 4, respectively. The boxes shown on the right side highlight the concept prototypes which contribute to the substantial differences between the two class-specific MCP distributions}
    \label{fig:fine_class_samples}
\end{figure*}

\begin{figure*}
    \centering
    \includegraphics[width=1.0\textwidth]{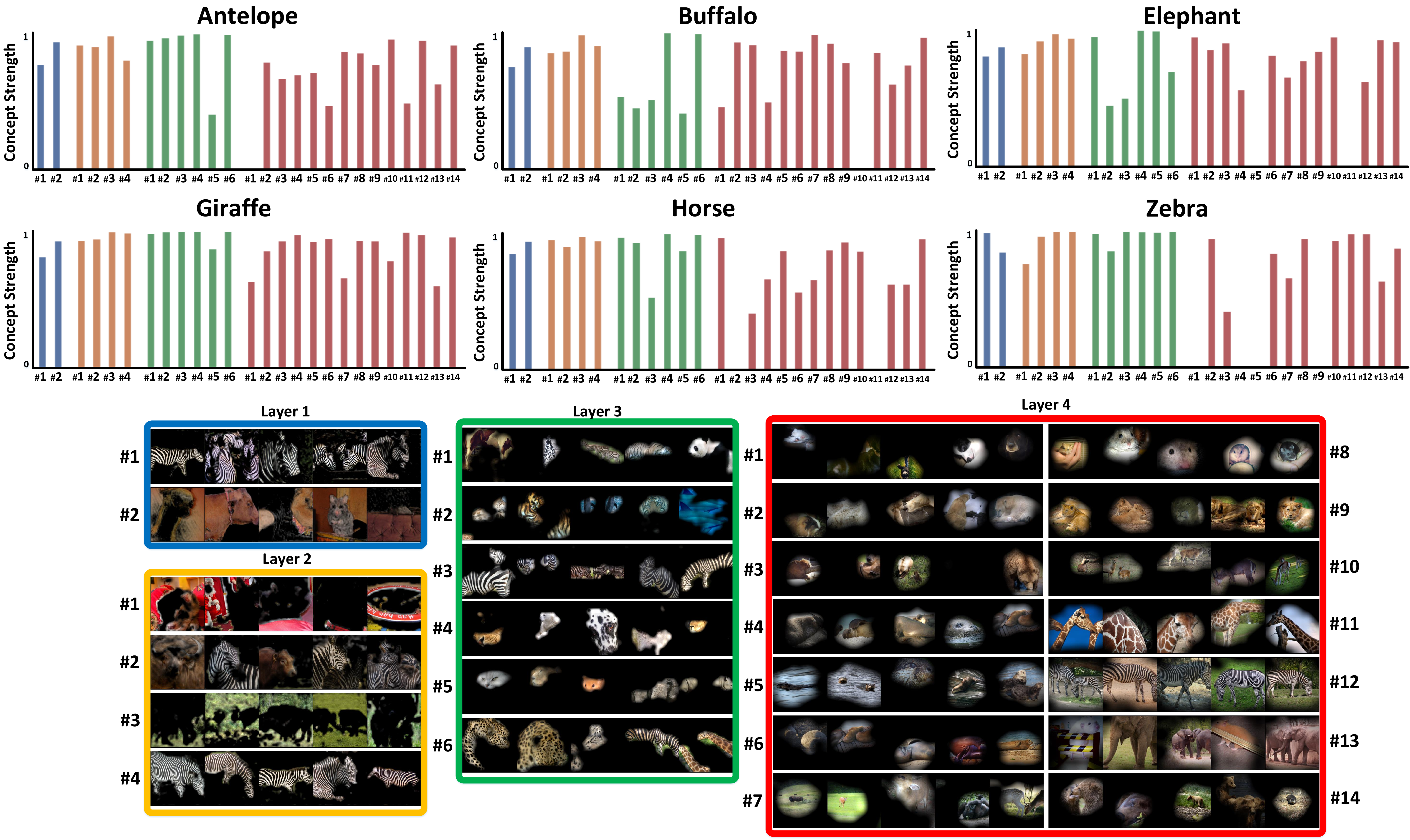}
    \caption{Class-wise explanation. For each category, the explanation is derived by aggregating the MCP distributions for the images of the target class. The concept strength indicates whether the concept prototypes are prevalent in most images belonging to the same class in the dataset. 
    }
    \label{fig:class_explanation}
\end{figure*}

\subsection{The effect of CKA loss}
To assess the disentangling effect of the Centered Kernel Alignment (CKA) loss, we observe that, despite a slight performance degradation when combined with the Class-aware Concept Distribution (CCD) loss, the disentanglement among concept prototypes becomes more significant due to the presence of the CKA loss, as illustrated in \Cref{fig:ablation_sample}. In the absence of the CKA loss, prototypes such as $\#1$ and $\#5$ in layer 2 seem to repeatedly learn the meaning of edges, while prototypes like $\#4$ and $\#5$ in layer 3 repeatedly focus on the bird's head with white color (i.e. there are prototypes being less disentangled).

\section{More explanation samples}

\subsection{Image-level explanations}
In \Cref{fig:instance_samples}, we illustrate how MCPNet classifies an image by the MCP distribution. We compute the MCP distribution for each image, reflecting the presence of concept prototypes. By aligning the image's MCP distribution with the class-specific MCP distribution, which signifies the strength of concept prototypes across the majority of images in the same class, the image is classified into the category most resembling its concept prototype strength.
Beyond image classification, the MCP distribution is also employed to generate explanations for individual instances, as depicted in \Cref{fig:coarse_samples}. Each computed concept prototype strength signifies the identified concept prototype on the image. Analyzing the strength of each concept prototype unveils how the model arrives at the classification decision for each image.

\begin{figure*}
    \centering
    \includegraphics[width=1.0\textwidth]{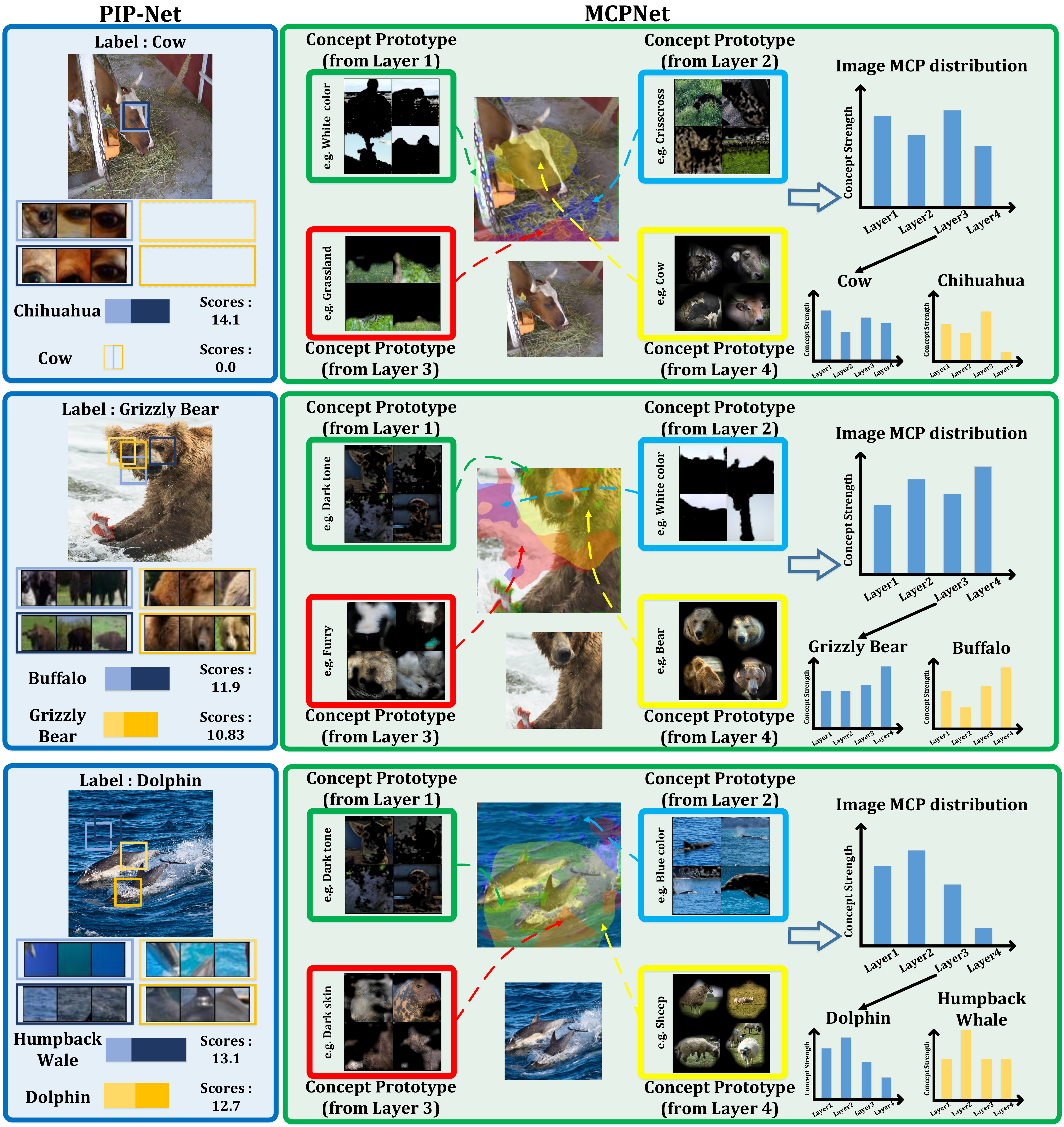}
    \caption{\bor{More explanation comparisons between MCPNet and PIP-Net~\cite{nauta2023pip}. PIP-Net~\cite{nauta2023pip} approaches all three scenarios with object-focused explanations, which results in inaccurate classifications. In the first scenario, PIP-Net fails to identify any matching prototypes from the correct class within the image. For the second and third scenarios, despite PIP-Net recognizing concepts from the correct class in the image, concepts from other classes receive higher responses, thus leading to confusion.
    Conversely, MCPNet employs multi-scale concept explanations as the foundation for accurate classification. In particular, for the second scenario, the high responses to both Grizzly Bear and buffalo classes in terms of high-level concept would lead to confusion if the classification is based solely on the high-level responses, while such confusion can be resolved with the incorporation of low-level concept responses. Moreover, in the third scenario, even without a direct concept match in the image -- such as the concept from layer 4, potentially interpreted as sheep -- MCPNet accurately interprets the image using the constructed MCP distribution based on the holistic consideration over the distribution of concept responses across multiple scales.}}
    
    \label{fig:more_pip_compare}
\end{figure*}

\bor{Several example results of further comparisons with PIP-Net~\cite{nauta2023pip} in terms of explanations are shown in \Cref{fig:more_pip_compare}, where PIP-Net~\cite{nauta2023pip} produces object-centric explanations thus leading to misclassification when images lack the object-centric prototypes related to the correct class or have confusion across object-centric prototypes from various classes. Conversely, MCPNet bases its classification on a multi-level concept response that incorporates both high-level and low-level concepts, in which such approach ensures that the entire spectrum of concept responses contributes to the variation in the MCP distribution.}

\subsection{Class-level explanations}
We will delve deeper into how our Multi-Level Concept Prototype Classifier (MCPNet) provides class-level explanations of the model through class-specific MCP distributions. The class-specific MCP distribution summarizes the distribution of concept prototype strengths within images of the same class, while our Class-aware Concept Distribution (CCD) loss also encourages the difference in such MCP distributions across classes.
The class-wise explanations are derived by examining the high or low responses of the concept prototype in class-specific MCP distribution, which indicate the presence or absence of specific concept prototypes for the corresponding class, as illustrated in \Cref{fig:class_explanation}.

We also illustrate the distinctions between the two MCP distributions for the fine-grained dataset and showcase the concept prototypes contributing to the differences between these distributions, as depicted in \Cref{fig:fine_class_samples}.

\section{Concept prototype importance and discriminativeness}
\begin{figure*}
    \centering
    \includegraphics[width=0.6\textwidth]{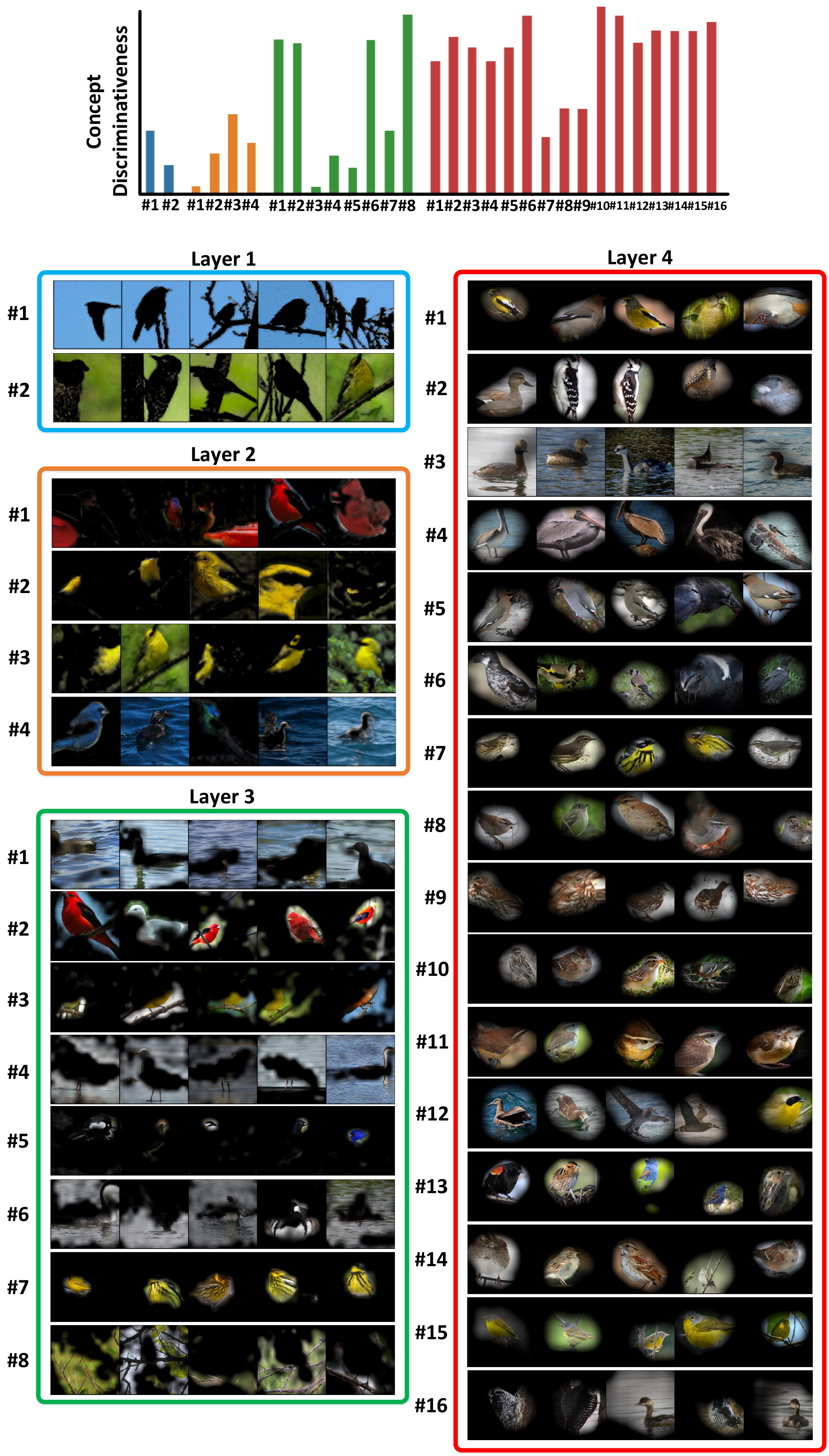}
    \caption{Concept discriminativeness at the class level is depicted in the upper part of the figure, showcasing the standard deviation of concept strength across classes. The higher variance in concept strength signifies that the corresponding concept prototypes play a more discriminative role in distinguishing images.}
    \label{fig:concept_imp}
\end{figure*}

Our MCPNet excels at revealing the importance or discriminativeness of each concept prototype, both at the image and class levels. For the image level, the importance of a concept prototype is directly inferred from the concept's strength, demonstrating its impact on the final prediction of classification. For the class level, we calculate the standard deviation of the strength for each concept within the class-specific MCP distribution to assess its discriminativeness across classes, as shown in \Cref{fig:concept_imp}. 

\clearpage

\clearpage
{
    \small
    \bibliographystyle{ieeenat_fullname}
    \bibliography{main}
}

\end{document}